%% file: main.tex
\documentclass[letterpaper, 10 pt, conference]{ieeeconf}  

\IEEEoverridecommandlockouts                              

\usepackage{multicol}
\usepackage{multirow}
\usepackage[usenames,dvipsnames]{xcolor}
\usepackage[colorlinks=true, citecolor=blue, linkcolor=black, urlcolor=blue]{hyperref}

\usepackage{times} 

\usepackage{graphicx}
\usepackage{algorithm}
\usepackage{algpseudocode}
\usepackage{amsmath}
\usepackage{amssymb}  
\usepackage{pifont}
\usepackage{nicefrac}       
\usepackage[normalem]{ulem}
\usepackage{capt-of}
\usepackage{booktabs}

\title{\LARGE \bf
\input{includes/meta/title}
}

\input{includes/meta/authors}

\begin{document}

\makeatletter
\let\@oldmaketitle\@maketitle
\renewcommand{\@maketitle}{\@oldmaketitle
\centering
\includegraphics[width=.96\linewidth]{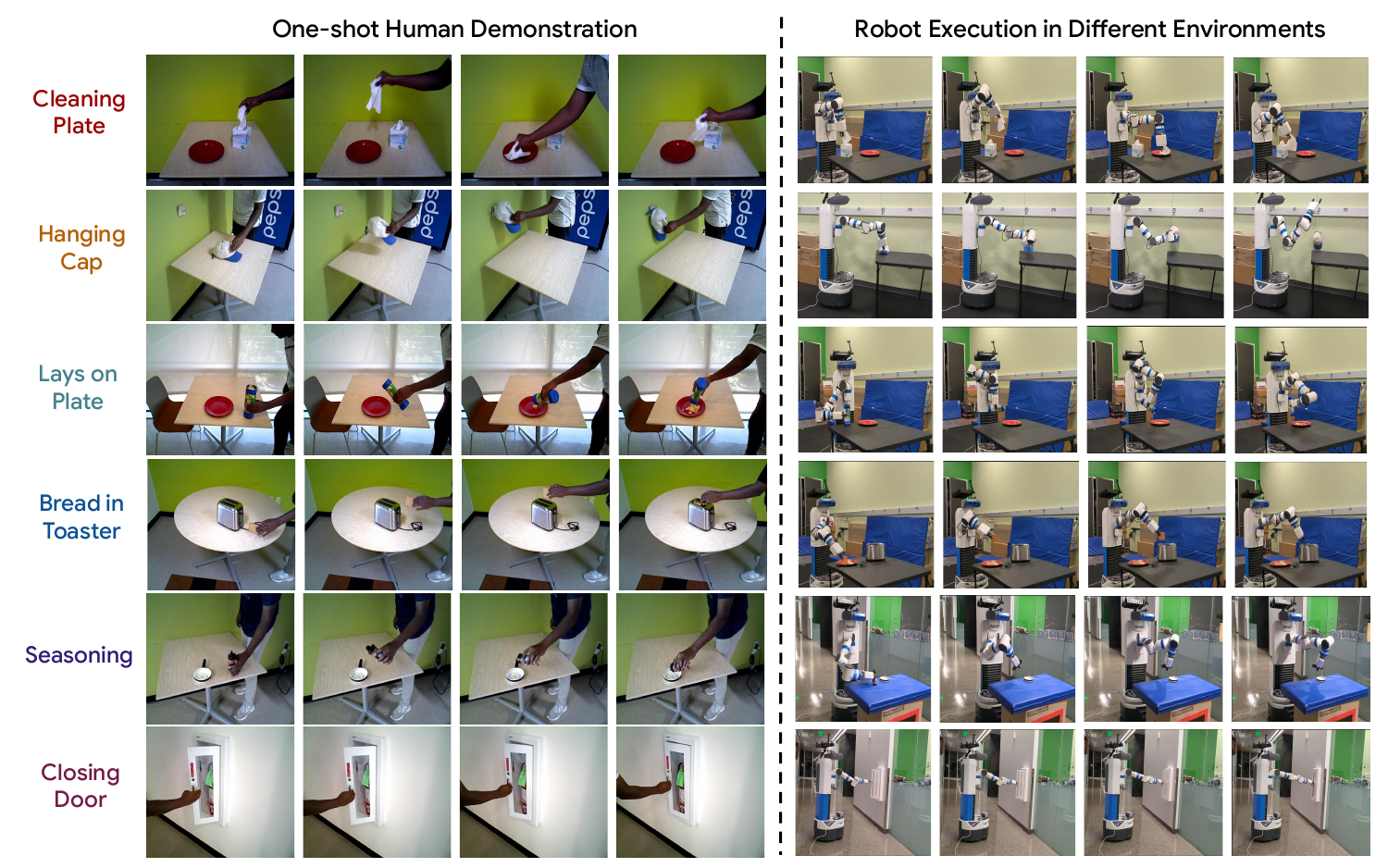}
\captionof{figure}{
Illustrations of several tasks that our system enables a mobile robot to perform.
}\label{fig:intro}
}
\makeatother

\maketitle
\thispagestyle{empty}
\pagestyle{empty}


\begin{abstract}
\input{includes/sections/0-abstract}
\end{abstract}

\input{includes/sections/1-intro}
\input{includes/sections/2-related-works}

\input{includes/sections/3-method}
\input{includes/sections/4-experiments}

\input{includes/sections/5-conclusion}
\input{includes/sections/6-limitations_and_futurework}

\section*{ACKNOWLEDGMENT}
This work was supported in part by the National Science Foundation (NSF) under Grant Nos. 2346528 and 2520553, and the NVIDIA Academic Grant Program Award.

\bibliographystyle{ieeetr}
\bibliography{references}

\end{document}

%% file: includes/meta/title.tex
HRT1: One-Shot Human-to-Robot Trajectory Transfer for \\  Mobile Manipulation


%% file: includes/meta/authors.tex
\author{Sai Haneesh Allu$^{*1}$, Jishnu Jaykumar P$^{*1}$, Ninad Khargonkar$^{1}$, Tyler Summers$^{1}$, Jian Yao$^{2}$, Yu Xiang$^{1}$
\thanks{*Equal contribution}
\thanks{$^{1}$The authors are with the University of Texas at Dallas, Richardson, TX 75080, USA {\tt\small \{SaiHaneesh.Allu, Jishnu.P, ninadarun.khargonkar, Tyler.Summers, yu.xiang \}@utdallas.edu}}%
\thanks{$^{2}$Jian Yao is with XPeng, Santa Clara, CA 95054, USA
        {\tt\small jian.yao@xiaopeng.com}}%
}

%% file: includes/sections/0-abstract.tex
We introduce a novel system for human-to-robot trajectory transfer that enables robots to manipulate objects by learning from human demonstration videos. The system consists of four modules. The first module is a data collection module that is designed to collect human demonstration videos from the point of view of a robot using an AR headset. The second module is a video understanding module that detects objects and extracts 3D human-hand trajectories from demonstration videos. The third module transfers a human-hand trajectory into a reference trajectory of a robot end-effector in 3D space. The last module utilizes a trajectory optimization algorithm to solve a trajectory in the robot configuration space that can follow the end-effector trajectory transferred from the human demonstration. Consequently, these modules enable a robot to watch a human demonstration video once and then repeat the same mobile manipulation task in different environments, even when objects are placed differently from the demonstrations. Experiments of different manipulation tasks are conducted on a mobile manipulator to verify the effectiveness of our system\footnote{Project page: \url{https://irvlutd.github.io/HRT1}}.

%% file: includes/sections/1-intro.tex
\section{Introduction}\label{sec: intro}

Building autonomous robots that can help people perform various tasks is the dream of every roboticist. To achieve this goal, we need to enable robots to manipulate objects. Traditionally, roboticists built manipulation systems by integrating perception, planning, and control. Recently, the data-driven learning-based paradigm has received more attention~\cite{chi2023diffusion,wen2023any,ze20243d,fu2024mobile,di2024dinobot}, where robot demonstrations are collected by teleoperation~\cite{fu2024mobile,he2024learning,padalkar2023open} to learn manipulation control policies. Although some recent efforts are devoted to collect large-scale demonstration datasets such as the Open X-Embodiment dataset~\cite{padalkar2023open} and the DROID dataset~\cite{khazatsky2024droid}, the scales of these robot datasets are still not large enough to learn generalizable manipulation skills. The high cost of teleoperation makes it impractical to create robot datasets on the scale of images or language.

On the other hand, human demonstration data is much easier to obtain compared to robot teleoperation. Researchers started exploring using human demonstration videos to teach robot manipulation. These human videos can come from the Internet~\cite{bharadhwaj2023zero,shaw2023videodex,bharadhwaj2024track2act}, motion capture~\cite{liu2025dextrack,li2025maniptrans,yuan2025hermes}, or simple camera setups~\cite{heppert2024ditto,li2024okami,kareer2024egomimic,lum2025crossing}, and different ways of using these human data are explored in these methods.

One popular way is co-training where human data and robot data are jointly used for learning control policies such as in EgoMimic~\cite{kareer2024egomimic} and MotionTracks~\cite{ren2025motion}. However, robot teleoperation data are still needed to make this approach work. For methods that only use human data, we can classify these into three categories. 1) Methods based on imitation learning transform human actions into robot actions for imitation learning~\cite{qin2022dexmv,fu2024humanplus,lepert2025masquerade}; 2) Reinforcement learning methods use human demonstrations in simulation to learn RL policies, where the reward function is designed based on the similarity between robot behavior and human behavior~\cite{liu2025dextrack,lum2025crossing,yuan2025hermes}; 3) Training-free methods transfer the human hand trajectories into robot motion trajectories using object motion~\cite{di2024dinobot,heppert2024ditto} or retargeting~\cite{li2024okami}. These approaches have their own pros and cons. For example, imitation learning methods typically require a number of human demonstrations for each task, which cannot be applied to a one-shot human demonstration. RL methods require building a digital twin of the task space, which is a challenging perception task. Training-free methods are highly dependent on the accuracy of the perception model for trajectory transfer.

In this work, we introduce a novel system, HRT1, for one-shot human-to-robot trajectory transfer that belongs to the training-free category mentioned above. The system consists of four modules designed to overcome the limitations of previous training-free methods~\cite{heppert2024ditto,li2024okami}. First, our data collection module uses an AR headset to capture human demonstration videos, where we can make sure that human hands are largely visible for recognition. Second, the video understanding module utilizes GroundingDINO~\cite{GDINO} for object detection, SAMv2~\cite{ravi2024sam2} for obtaining object masks, and HaMeR~\cite{pavlakos2024HaMeR} for hand pose estimation. It also uses depth images to compute accuracy 3D hand poses. Third, we utilize the unified gripper coordinate space introduced in~\cite{khargonkar2024robotfingerprint} to transfer a 3D hand pose to a robot gripper pose. Consequently, we obtain a trajectory of the robot gripper from the human demonstration video. Finally, in order to transfer this gripper trajectory to a new task space, we carefully leverage BundleSDF~\cite{Wen2023-bundleSDF} to estimate the pose transformations between objects in the demonstration video and in the task space, and then apply these transformations to the demonstration trajectory. More importantly, we have designed a new trajectory optimization algorithm to solve the robot configurations that follow the transformed gripper trajectory in the task space. We can also optimize the robot base position that reaches the transformed gripper trajectory. Overall, the system enables a robot to watch a human demonstration video once, and repeat the same mobile manipulation task in new environments with different object arrangements.

Compared to DITTO~\cite{heppert2024ditto} which transfers object pose trajectories, our system transfers human hand trajectories. During manipulation tasks, objects are usually occluded by human hands. We found that DITTO cannot obtain reliable object poses when the manipulated objects are largely occluded. Compared to OKAMI~\cite{li2024okami} that simply replies on retargeting and inverse kinematics, our trajectory optimization algorithm can deal with noises in the transferred gripper trajectory. In addition, our system is the only training-free method that supports mobile manipulation.

We demonstrate the effectiveness of HRT1 in 16 diverse tasks, including grasping household objects, manipulating tools, and performing fine motor actions. Our results show that HRT1 achieves promising performance in one-shot manipulation scenarios. It significantly outperforms DITTO~\cite{heppert2024ditto} in these tasks. Fig.~\ref{fig:intro} illustrates several tasks that our system enables a Fetch mobile manipulator to perform.

%% file: includes/sections/2-related-works.tex
\section{Related Work}\label{sec: label}

\begin{table*}[]
\resizebox{0.95\linewidth}{!}{%
\begin{tabular}{c|c|c|c|c|c|c|c}
\hline
\textbf{Method} & \textbf{Year} &
  \textbf{Data} &
  \textbf{Motion Generation} &
  \textbf{Representation} &
  \textbf{Mobile Manipulation} &
  \textbf{1-Shot Imitation} &
  \textbf{Training Free} \\ \hline

DexMV~\cite{qin2022dexmv} & 2022 & Human & Imitation Learning & Hand-Object Pose  & \textcolor{red}{\ding{55}} & \textcolor{red}{\ding{55}} & \textcolor{red}{\ding{55}}  \\ \hline

DOME~\cite{valassakis2022demonstrate} & 2022 & Robot & Replay Demonstration & Object Appearance &  \textcolor{red}{\ding{55}} & \textcolor{ForestGreen}{\ding{51}} & \textcolor{red}{\ding{55}} \\ \hline

Trajectory Transfer~\cite{vitiello2023one} & 2023 & Robot & Trajectory Transfer & Object Pose &  \textcolor{red}{\ding{55}} & \textcolor{ForestGreen}{\ding{51}} & \textcolor{red}{\ding{55}} \\ \hline

DINOBot~\cite{di2024dinobot} & 2024 & Robot & Replay Demonstration & Object Appearance & \textcolor{red}{\ding{55}} & \textcolor{ForestGreen}{\ding{51}} & \textcolor{ForestGreen}{\ding{51}}  \\ \hline

ScrewMimic~\cite{bahety2024screwmimic} & 2024 & Human & Imitation Learning & Hand Pose & \textcolor{red}{\ding{55}} & \textcolor{ForestGreen}{\ding{51}} & \textcolor{red}{\ding{55}} \\ \hline
  
DITTO~\cite{heppert2024ditto} & 2024 & Human & Inverse Kinematics & Object Pose & \textcolor{red}{\ding{55}} & \textcolor{red}{\ding{55}} & \textcolor{ForestGreen}{\ding{51}} \\ \hline

HumanPlus~\cite{fu2024humanplus} & 2024 & Human & Imitation Learning & Hand Pose &  \textcolor{red}{\ding{55}} & \textcolor{red}{\ding{55}} & \textcolor{red}{\ding{55}} \\ \hline

OKAMI~\cite{li2024okami} & 2024 & Human & Inverse Kinematics & Hand Pose & \textcolor{red}{\ding{55}} & \textcolor{ForestGreen}{\ding{51}} & \textcolor{ForestGreen}{\ding{51}}  \\ \hline

ATM~\cite{wen2023any} & 2024 & Human+Robot & Imitation Learning & Point Trajectory & \textcolor{red}{\ding{55}} & \textcolor{red}{\ding{55}} & \textcolor{red}{\ding{55}} \\ \hline

EgoMimic~\cite{kareer2024egomimic} & 2024 & Human+Robot & Imitation Learning & Hand Pose & \textcolor{red}{\ding{55}} &  \textcolor{red}{\ding{55}} & \textcolor{red}{\ding{55}}  \\ \hline

HUMAN2SIM2ROBOT~\cite{lum2025crossing} & 2025 & Human & Reinforcement Learning & Hand-Object Pose & \textcolor{red}{\ding{55}} & \textcolor{ForestGreen}{\ding{51}} & \textcolor{red}{\ding{55}} \\ \hline

MotionTracks~\cite{ren2025motion} & 2025 & Human+Robot & Imitation Learning & Hand Keypoint & \textcolor{red}{\ding{55}} &  \textcolor{red}{\ding{55}} & \textcolor{red}{\ding{55}}  \\ \hline

AMPLIFY~\cite{collins2025amplify} & 2025 & Human+Robot & Imitation Learning & Point Trajectory & \textcolor{red}{\ding{55}} & \textcolor{red}{\ding{55}} & \textcolor{red}{\ding{55}} \\ \hline

HERMES~\cite{yuan2025hermes} & 2025 & Human & Reinforcement Learning & Hand-Object Pose & \textcolor{ForestGreen}{\ding{51}} & \textcolor{ForestGreen}{\ding{51}} & \textcolor{red}{\ding{55}} \\ \hline

Masquerade~\cite{lepert2025masquerade} & 2025 & Human+Robot & Imitation Learning & Hand Pose & \textcolor{red}{\ding{55}} & \textcolor{red}{\ding{55}} & \textcolor{red}{\ding{55}} \\ \hline

ImMimic~\cite{liu2025immimic} & 2025 & Human+Robot & Imitation Learning & Hand Pose &  \textcolor{red}{\ding{55}} & \textcolor{red}{\ding{55}} & \textcolor{red}{\ding{55}} \\ \hline

EMMA~\cite{zhu2025emma} & 2025 & Human+Robot & Imitation Learning & Hand Pose & \textcolor{ForestGreen}{\ding{51}} & \textcolor{red}{\ding{55}} & \textcolor{red}{\ding{55}} \\ \hline

\textbf{HRT1 (Ours)} & 2025 & Human & Trajectory Optimization & Hand Pose & \textcolor{ForestGreen}{\ding{51}}   &  \textcolor{ForestGreen}{\ding{51}} & \textcolor{ForestGreen}{\ding{51}}    \\ \hline
\end{tabular}%
}
\caption{Comparison between existing methods on robot manipulation using human and/or robot demonstrations.}
\label{tab:related_world}
\vspace{-4mm}
\end{table*}

Our work is related to one-shot imitation learning methods for robot manipulation in the literature. One paradigm to tackle the one-shot imitation learning problem is via meta-learning~\cite{duan2017one,finn2017one,yu2018one,wu2024semi}, where demonstrations of various tasks are available, and meta-learning is applied to learn policies that can utilize one-shot demonstration of new tasks during testing. The main limitation of meta-learning-based imitation learning methods is that their generalizability to new tasks is limited. These methods usually assume that there is some similarity between test tasks and training tasks in order to train a policy that can leverage the similarity between tasks. If a new task is very different from the training tasks, the policy may fail. Another limitation is that these methods require a large number of demonstrations of training tasks to learn the policy.

Instead of learning manipulation policies that can be adapted to new tasks, another paradigm for one-shot imitation learning focuses on trajectory transfer using one-shot demonstrations. Given a one-shot demonstration of a new task, methods in this paradigm transfer the motion trajectory in the demonstration to a robot. We classify these trajectory transfer methods into the following two categories.

\textbf{Object-Centric Trajectory Transfer.} Methods in this category first extract object trajectory in one-shot demonstrations and then design algorithms to allow a robot to repeat the extracted object trajectory. For example, \cite{lee2015non,lee2015learning,schulman2016learning} utilized non-rigid registration of deformable objects for trajectory transfer. \cite{wen2022you} utilized category-level object pose tracking to extract object trajectories from human demonstration videos and then used robot motion planning and behavior cloning to transfer the trajectories. \cite{vitiello2023one} utilized unseen object pose estimation to transfer end-effector trajectories of a robot to new configurations of objects. Similarly, DITTO~\cite{heppert2024ditto} also used object pose estimation with feature correspondences for trajectory transfer. DOME~\cite{valassakis2022demonstrate} and DINOBot~\cite{di2024dinobot} learned a visual servo network to align the end-effector poses between testing and demonstration using object images. Object-centric methods benefit from state-of-the-art object perception algorithms for object tracking. However, since the robot gripper or human hands are not used in the demonstration, these methods usually require external algorithms for grasp planning and motion planning. Another limitation is that object tracking and matching may fail due to occlusions between hands and objects. Consequently, it is challenging to due with small objects for object-centric methods.

\textbf{Human-Centric Trajectory Transfer.} Human-centric methods for trajectory transfer try to understand how human hands manipulate objects. These methods extract human hand trajectories from demonstration videos and design algorithms to transfer the extracted trajectories to a robot~\cite{li2021meta,xiong2021learning}. For example, DexMV~\cite{qin2022dexmv} uses hand pose estimation and object pose estimation to extract the hand-object trajectories from videos followed by imitation learning to learn a manipulation policy. ScrewMimic~\cite{bahety2024screwmimic} extracts the screw motion between hands and objects, and then learns a policy to transfer the screw actions. HumanPlus~\cite{fu2024humanplus} leverages human pose estimation to re-target the human motion to a humanoid. A similar idea is also explored in OKAMI~\cite{li2024okami}.

These methods leverage state-of-the-art methods for human pose estimation such as SLAHMR~\cite{ye2023decoupling} and World-Grounded Humans with Accurate Motion (WHAM)~\cite{shin2024wham} and hand pose estimation such as HaMeR~\cite{pavlakos2024HaMeR} to understand human demonstration videos. They differ in how to transfer the human trajectories to a robot. Existing methods either learn some manipulation policies~\cite{li2021meta,xiong2021learning,qin2022dexmv,bahety2024screwmimic} which is very challenging in one-shot imitation settings or using simple motion re-targeting strategies~\cite{fu2024humanplus,li2024okami}. In our work, we utilize a recent grasp representation named the unified gripper coordinate space~\cite{khargonkar2024robotfingerprint} to transfer human hand poses to robot gripper poses, and then design a trajectory optimization algorithm to generate the robot motion. Since the trajectory optimization is robust to noises in the human demonstrations, our method improves over motion re-targeting alternatives. Table~\ref{tab:related_world} summarizes a number of recent works on using human demonstrations for robot manipulation.

%% file: includes/sections/3-method.tex
\section{HRT1 System}\label{sec: method}

\subsection{Human Demonstration Collection}\label{sub-sec:data-collection}

\addtocounter{figure}{-1}
\begin{figure}[h]
    \centering
    \includegraphics[width=\linewidth]{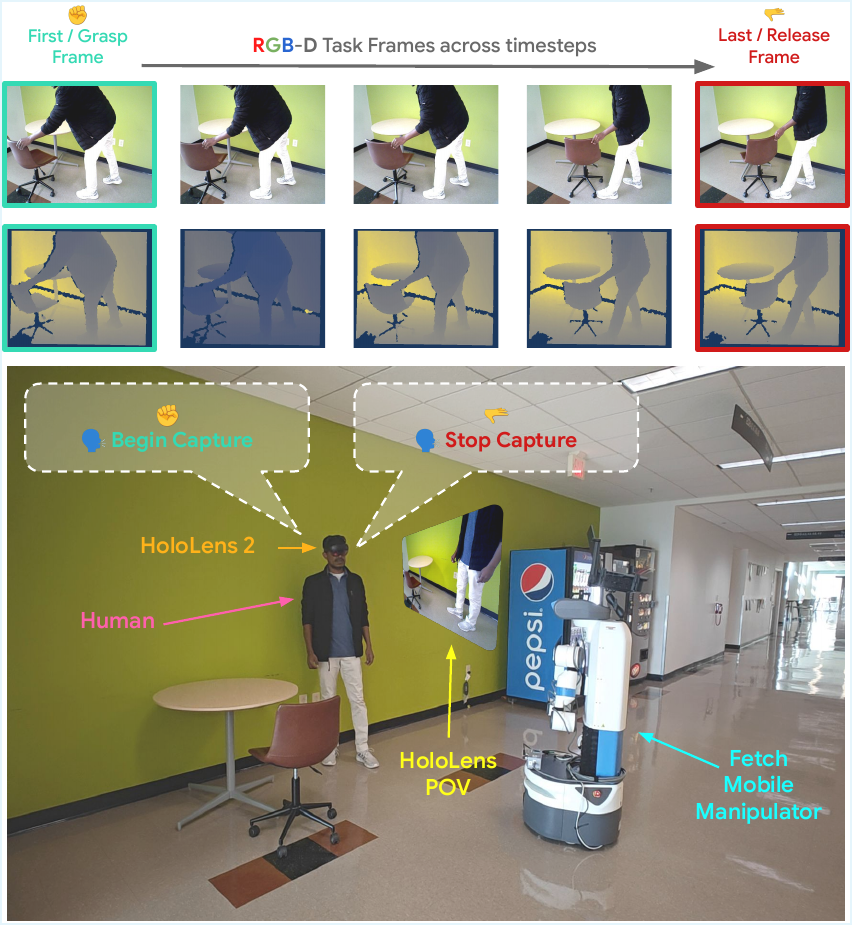}
    \vspace{-0.5cm}
    \caption{An example of a real-world scene where a human wearing a HoloLens2 device gives voice commands to capture RGB-D data for moving a chair. A Fetch mobile manipulator is used for the data capture process.}
    \label{fig:dc-pipeline}
\end{figure}

Building on the \textit{iTeach} framework~\cite{padalunkal2024iteach}, we have designed a data collection module that involves a human operator wearing a HoloLens 2 device and a Fetch mobile manipulator, as illustrated in Fig.~\ref{fig:dc-pipeline}. The HoloLens and the robot are connected through a Wi-Fi hotspot hosted by the robot, with the ROS server running on the robot. The robot continuously publishes its RGB camera feed to the ROS server, which the HoloLens subscribes to. \emph{This setup allows the HoloLens to display the point of view of the robot, enabling the user to visualize what the robot sees in real time.}


\begin{figure*}
    \centering
    \includegraphics[width=0.95\linewidth]{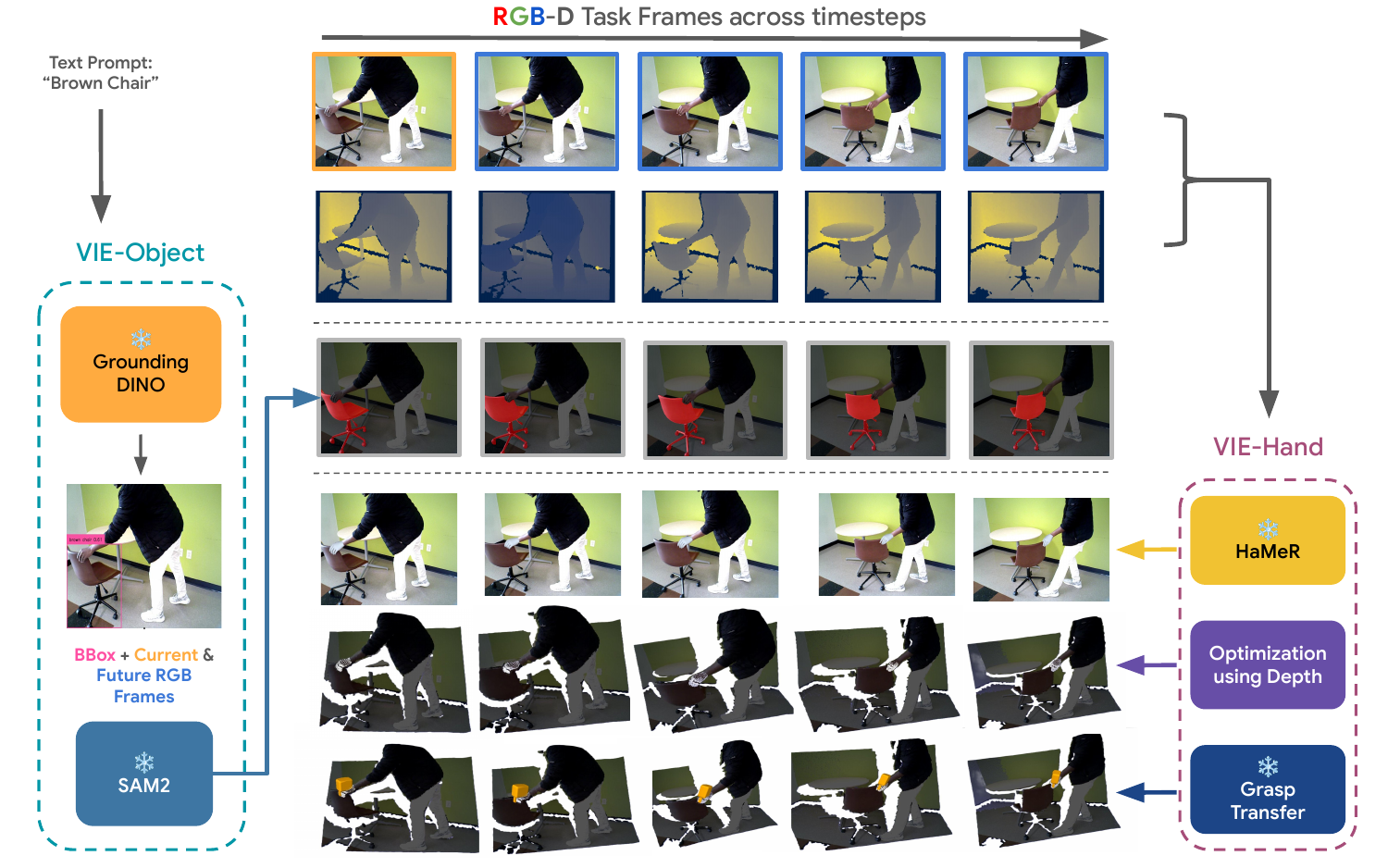}
    \vspace{-2mm}
    \caption{The overview of the Video Information Extraction (VIE) module for extracting object and hand information from a human demonstration video. }
    \label{fig:vie}
\end{figure*}

Using this capability, the operator can navigate the robot to ensure that the task space, their hands, and the objects of interest are all within view of the robot. Once the robot is placed for an optimal perspective, the user issues voice commands through the HoloLens, which are transmitted to the ROS server. For example, saying ``Stream'' displays the real-time RGB feed of the robot on the HoloLens. The command ``Begin capture'' initiates the recording of RGB-D data and camera poses on the robot file system. When the user says ``Stop capture'', the data recording is stopped for that specific demonstration.

The voice command-based interaction between the HoloLens and the user simplifies the demonstration process, allowing the operator to focus on robot navigation and task demonstration without distractions. The structured workflow of ``stream / begin capture / stop capture'' ensures efficient and rapid data collection for each task. This streamlined approach, combined with the focused scene composition, facilitates the accurate capture of interaction data in real time. To simplify the perception understanding of the human demonstration video, \emph{the operator carefully poses the hand gesture during manipulation similar to a two-finger robotic gripper.} In addition, we make sure that the first frame of the video starts with the grasping pose, and the last frame of the video corresponds to the grasp releasing pose, which can be easily achieved using voice commands for data capture.

\subsection{Demonstration Video Understanding}
\label{sub-sec:vie}
After data recording is complete, the next step is to extract object and hand information from task-specific video data. This module, referred to as the Video Information Extraction (VIE), is crucial for processing the recorded demonstrations.

\textbf{VIE-Object.} To obtain the segmentation masks of objects in the videos, we use the combination of GroundingDINO~\cite{GDINO} and SAM2~\cite{ravi2024sam2}. First, GroundingDINO~\cite{GDINO} is used with an appropriate object text prompt to detect the object and generate a bounding box $\mathcal{B}_1$ in the first frame, where the user provides the text prompt. The bounding box $\mathcal{B}_1$ is then provided as a prompt for SAM2~\cite{ravi2024sam2}, which generates object masks for subsequent frames. An illustration of the VIE-Object pipeline is shown in Fig.~\ref{fig:vie}.

\textbf{VIE-Hand.} A critical component of the VIE module is the recovery of accurate 3D hand poses from the video data, which is essential for transferring the human-hand trajectory to the robot to replicate the demonstrated task. We apply HaMeR~\cite{pavlakos2024HaMeR} on individual RGB frames for hand pose estimation. It outputs the $N$ vertices $\mathbf{V} = \{\mathbf{v}_1, \ldots, \mathbf{v}_N \}, \mathbf{v}_i \in \mathbb{R}^3, i=1,\ldots,N$ of the MANO~\cite{MANO:SIGGRAPHASIA:2017} hand mesh under the estimated 3D hand joints, and the 3D translation $\mathbf{t}_{\text{hamer}} \in \mathbb{R}^3$ of the hand. In addition, HaMeR assumes a virtual camera with the intrinsic matrix of the camera $\mathbf{K}_{\text{hamer}} \in \mathbb{R}^{3 \times 3}$. We noted that the 3D translation $\mathbf{t}_{\text{hamer}}$ from HaMeR~\cite{pavlakos2024HaMeR} is usually not accurate, which causes the MANO hand mesh to not align well with the 3D points of the hand from the RGB-D camera. This misalignment is largely due to a translation error along the $z$-axis of the camera since HaMeR only uses RGB images for hand pose estimation. To address this issue, we propose an optimization algorithm to improve the 3D translation $\mathbf{t}$.

First, we find the segmentation mask of the human hand in the image by projecting the estimated MANO hand mesh to the image using the estimated 3D translation $\mathbf{t}_{\text{hamer}}$ and the virtual camera intrinsics $\mathbf{K}_{\text{hamer}}$. Second, given the segmentation mask, we obtain the 3D point cloud of the hand by masking the point cloud from the RGB-D camera. Let us denote the point cloud of the hand by $\mathbf{P}_h$. Finally, we solve the following optimization problem to obtain a new 3D translation $\mathbf{t}^*$ for the MANO hand:
\begin{align}
    \mathbf{t}^* &= \arg \min_{\mathbf{t}} \Big( \lambda \underbrace{\sum_{i=1}^N \min_{\mathbf{p} \in \mathbf{P}_h} \| \mathbf{v}_i +\mathbf{t} - \mathbf{p} \|}_{\text{3D distance error}} +  \\   
    & \underbrace{\sum_{i=1}^N \| \text{proj}(\mathbf{K}_{\text{hamer}}, \mathbf{v}_i + \mathbf{t}_{\text{hamer}}) - \text{proj}(\mathbf{K}_{\text{real}}, \mathbf{v}_i +\mathbf{t}) \|^2}_{\text{2D projection error}} \Big),\nonumber
\end{align}
where we minimize the 3D distance error between the translated MANO vertices and the point cloud of the hand, and the 2D projection error between the projected points using the HaMeR translation $\mathbf{t}_{\text{hamer}}$ and the new translation $\mathbf{t}$. $\mathbf{K}_{\text{real}}$ indicates the intrinsics of the camera in the real world, and $\text{proj}()$ is the camera projection function using the camera intrinsics. $\lambda$ is a hyper-parameter to balance the two error terms. Fig.~\ref{fig:vie} illustrates the alignment between the MANO hand mesh and the point cloud of the human hand after translation optimization. We obtained accurate human-hand poses for most of the video frames.

\subsection{Human-to-Robot Grasp Transfer}\label{sub-sec:human-to-robot-hand}

\begin{figure}
    \centering
    \includegraphics[width=\linewidth]{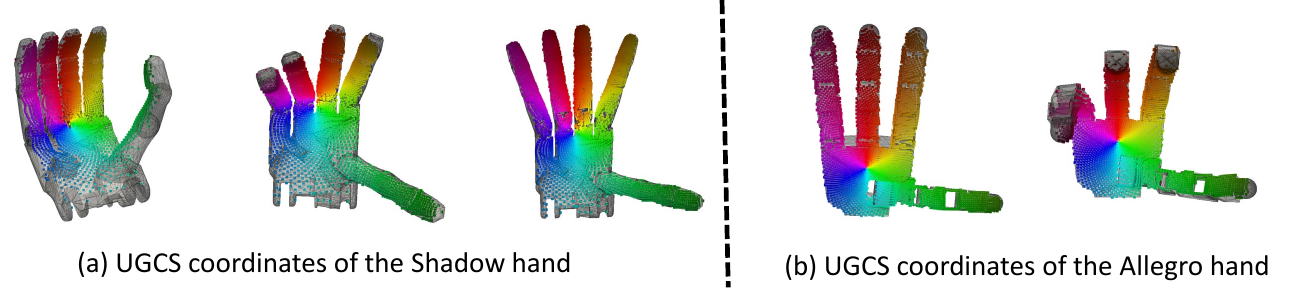}
    \vspace{-6mm}
    \caption{Illustration of the UGCS coordinates~\cite{khargonkar2024robotfingerprint} for the Shadow hand and the Allegro hand.}
    \label{fig:ugcs}
\end{figure}

Given the human hand poses from the VIE-Hand module, we aim to transfer the human grasps to the Fetch grasps. The objective of this module is to ensure that the transferred robot grasps are close to the reference human grasps in form and function. We use the unified gripper coordinate space (UGCS) representation from~\cite{khargonkar2024robotfingerprint} for grasp transfer, which enables grasp transfer between hands with different numbers of fingers and kinematics without manual re-targeting. Specifically, each gripper \(G\), is represented by a set of its interior surface points \(P_G = \{\mathbf{v}_g \; | \; \mathbf{v}_g \in \mathbb{R}^3 \}\) and an associated UGCS coordinate map \(\Phi_G\) with coordinates \( (\lambda, \varphi) \) for each point: 
\begin{equation}
    \Phi_{G} =  \{ (\lambda_{\mathbf{v}_g}, \varphi_{\mathbf{v}_g}) \; | \; \mathbf{v}_g \in P_G \; ; \; \lambda_{\mathbf{v}_g}, \varphi_{\mathbf{v}_g} \in [0, 1]  \},
\end{equation}
where $\lambda$ and $\varphi$ denote the normalized spherical coordinates of a surface point when the hand grasps a maximum sphere~\cite{khargonkar2024robotfingerprint}.

\begin{figure}
    \centering
    \includegraphics[width=\linewidth]{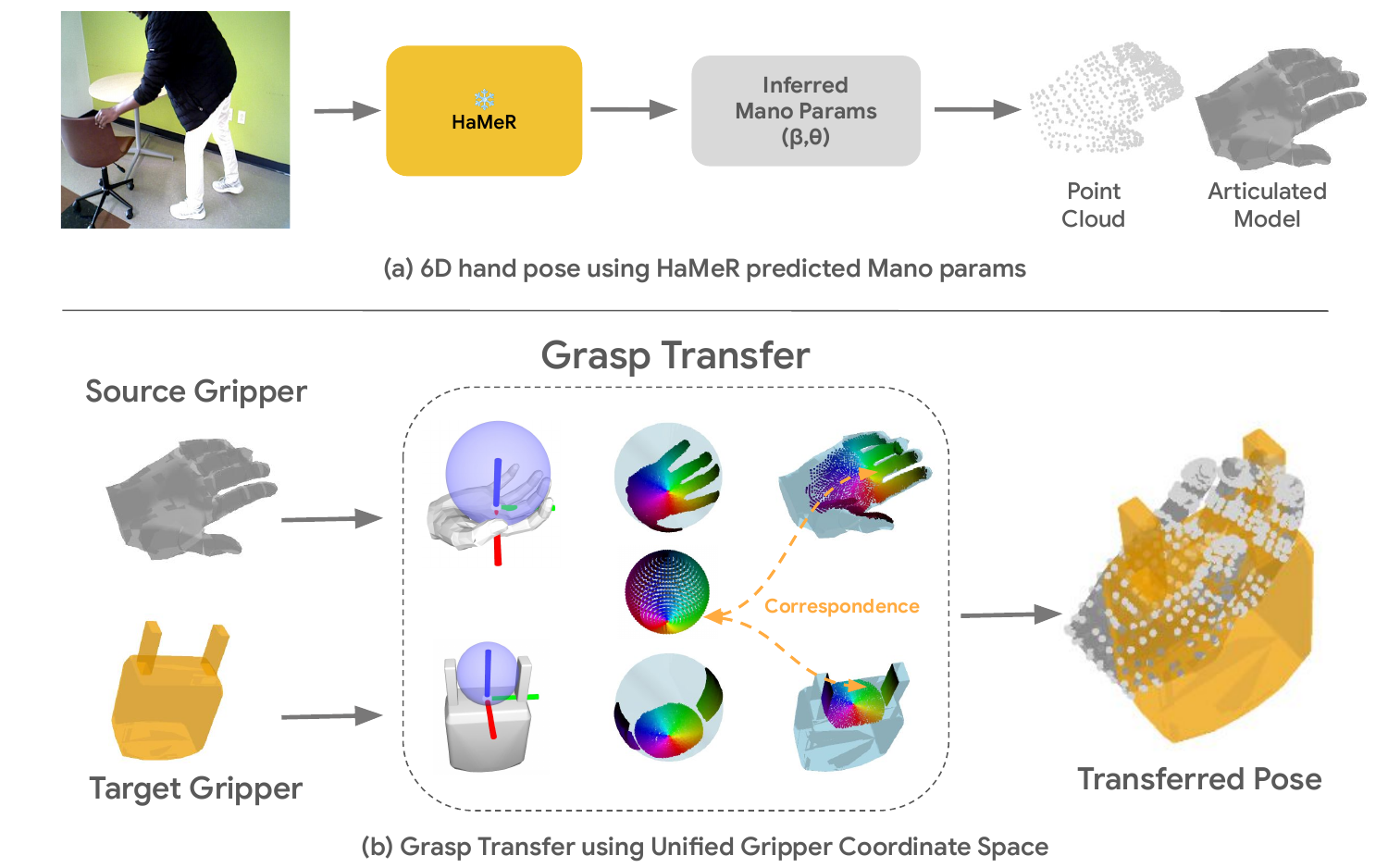}
    \vspace{-4mm}
    \caption{Illustration of the human hand to the Fetch gripper grasp transfer pipeline: (a) Inferring the 6D pose and joint values for an articulated hand model from HaMeR~\cite{pavlakos2024HaMeR}. (b) Grasp transfer: establishing correspondence between the human hand and the Fetch gripper points via the unified gripper coordinate space~\cite{khargonkar2024robotfingerprint}.}
    \label{fig:grasp_transfer}
    \vspace{-3mm}
\end{figure}

The cardinality of \(\Phi_G\) need not be the same for each gripper, and the gripper points \(\mathbf{v}_g\) change according to the grasp \(\mathbf{q}\), i.e., the configuration of the gripper. In other words, the gripper points \(\mathbf{v}_g\) change their 3D positions according to different configuration \(\mathbf{q}\). However, the UGCS coordinates of each point $(\lambda, \varphi)$ remain the same, which is independent of the configuration \(\mathbf{q}\). Fig.~\ref{fig:ugcs} illustrates the UGCS coordinates for the Shadow hand and the Allegro hand in different configurations.

A key advantage of the UGCS representation is that gripper points lying on functionally corresponding regions of two grippers will likely share similar UCGS coordinates \( (\lambda, \varphi) \). Therefore, we can establish a correspondence between the interior surface points of two grippers \(G_1, G_2\) by using their coordinate maps \(\Phi_{G_1}, \Phi_{G_2}\). Specifically, we first compute the pairwise distances between the coordinates in the two maps using the Haversine distance, since \(\lambda, \varphi\) are spherical coordinates. Next, we find mutually closest pairs across the two sets so that we have strong correspondence. This process establishes corresponding subsets of the two gripper surface points \(P_{G_1}^c \subset P_{G_1} \; ; \; P_{G_2}^c \subset P_{G_2}\) with \(|P_{G_1}^c| = |P_{G_2}^c|\) since we use the mutually closest pairs.  

Using the correspondence established by the UCGS coordinates across two grippers, we formulate the grasp transfer problem as an optimization problem.  Let the target grasp be represented as \(\mathbf{q}_F\) and the reference grasp as \(\mathbf{q}_H\). The desired target grasp \(\mathbf{q}_F\) is obtained by minimizing an objective function that considers the closeness between the corresponding source and target gripper points, and the validity of the joint value. A differentiable kinematics model is used to provide gradients in \(\mathbf{q}_F\) and we minimize the objective using the gradient descent with Adam optimizer~\cite{diederik2014-adam-optimizer}. The optimization problem is defined as
\begin{equation}
\mathbf{q}_F^* = \arg\min_{\mathbf{q}_F} \;  E_{\text{dist}}(P_{H}^c(\mathbf{q}_H), P_{F}^c (\mathbf{q}_F)) + E_{\text{n}}(\mathbf{q}_F),
\end{equation}
where \(P_{H}\) is the fixed set of human hand points with the fixed reference human grasp \(\mathbf{q}_H\). \(P_{F}(\mathbf{q}_F)\) is the set of Fetch gripper points dependent on the current grasp \(\mathbf{q}_F\). Once $\mathbf{q}_F$ is updated, we recompute the gripper points $P_{F}$ using forward kinematics.
Similarly, \(P_H^c, P_F^c\) are the corresponding subsets of human and Fetch gripper points, respectively. We only need to establish the correspondence once, before starting the optimization, since even though the points may change, the subset indexing order remains fixed. The objective function for optimization involves terms: 1) \(E_{\text{dist}}\) computes the mean Euclidean distance between corresponding gripper points; 2) $E_{\text{n}}$ from GenDexGrasp~\cite{li2022gendexgrasp} penalizes the grasp configuration \(\mathbf{q}_F\) if it is outside the joint limits. Since the Fetch gripper is a two-finger gripper, the grasp configuration $\mathbf{q}_F$ is a homogeneous transformation $\mathbf{T} \in \mathbb{SE}(3)$. Fig.~\ref{fig:grasp_transfer} illustrates the transfer from human grasps to Fetch gripper grasps.

Given a human hand trajectory from a human demonstration, we perform the above grasp transfer offline for each human hand grasp, which outputs a trajectory $\mathcal{T}^{\text{demo}}= \left\{ \mathbf{T}_{i}^{\text{demo}} \right\}_{i=1}^T$ of Fetch gripper grasps, as illustrated in the last row of Fig.~\ref{fig:vie}, where $T$ is the number of grasps.


\subsection{Trajectory Alignment for Task Execution}\label{sub-sec:traj-in-obj-frame}

\begin{figure}[!h]
    \centering
    \includegraphics[width=\linewidth]{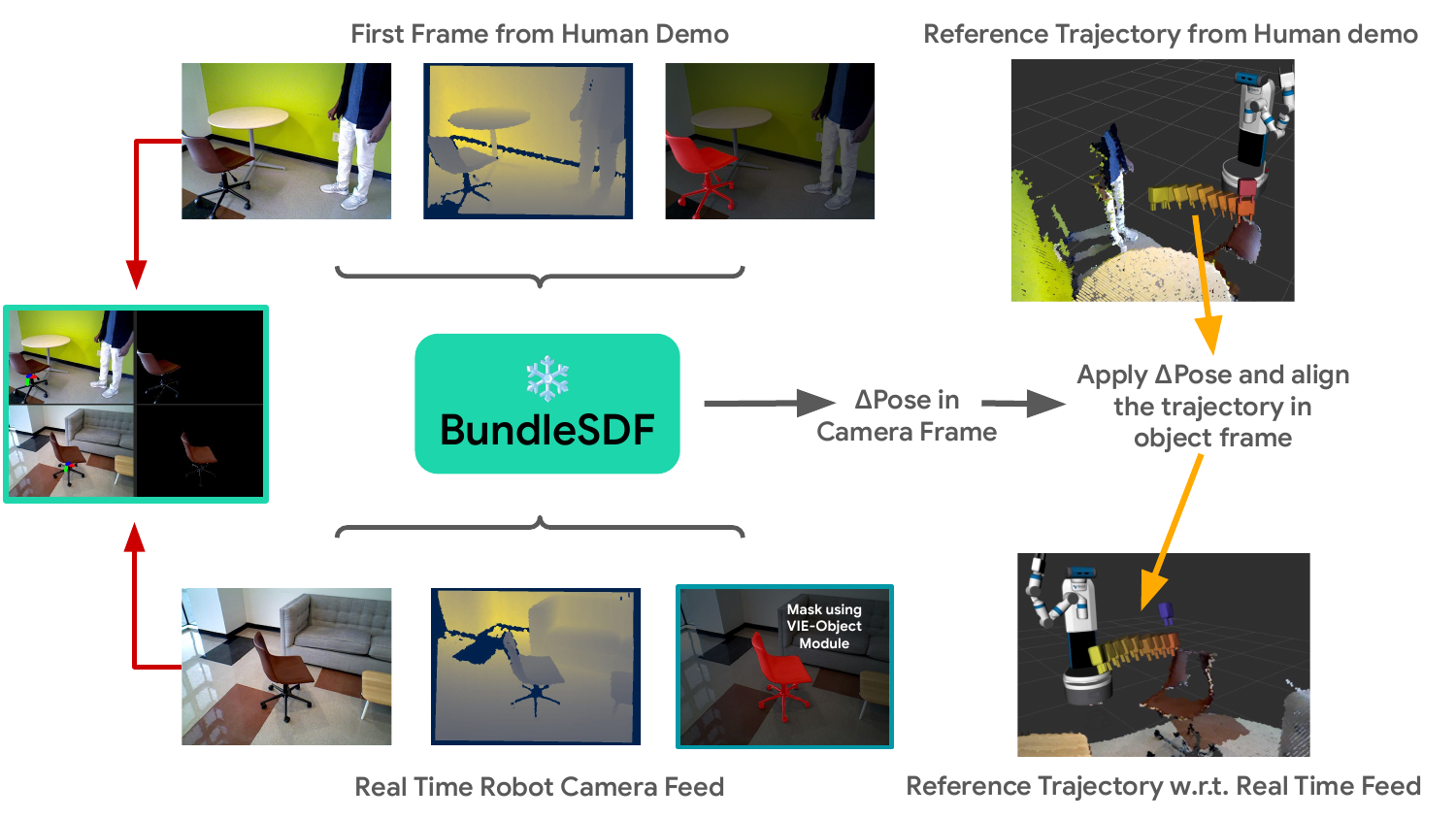}
    \vspace{-6mm}
    \caption{An overview of the trajectory alignment pipeline within an object-centric reference frame using BundleSDF~\cite{Wen2023-bundleSDF}. This module is specifically designed to preserve consistency in demonstrations across diverse scene configurations and varying object poses.}
    \label{fig:traj-in-obj-frame}
\end{figure}

In real-world scenarios, we hope that the robot can repeat the demonstrated tasks in different environments. Therefore, it is necessary to represent the transferred trajectory $\mathcal{T}^{\text{demo}}$ in an object-centric reference frame rather than in a fixed global frame. In our system, we align the trajectory according to the pose of the object, which ensures that the transferred trajectory remains consistent during the execution of the task. 

\textbf{Notations.} We adopt the notation $\mathbf{T}_{ba}$ to represent the pose of frame $\{a\}$ with respect to frame $\{b\}$.  $\mathbf{T}_{ba} = ({\mathbf{R}}_{ba}, {\mathbf{t}}_{ba}) \in \mathbb{SE}(3)$, where ${\mathbf{R}}_{ba}$ and ${\mathbf{t}}_{ba}$ are the 3D rotation and 3D translation of frame $\{a\}$ with respect to frame $\{b\}$, respectively. The important frames in this work are frame $\{b\}$ for the base of the robot, frame $\{c\}$ for the camera of the robot, frame $\{o\}$ or $\{o_1\}$ for the manipulated object, frame $\{o_2\}$ for the secondary object and frame $\{e\}$ for the end effector of the robot. 

Using the convention above, the transferred trajectory of the robot end-effector $\mathcal{T}^{\text{demo}}= \left\{ \mathbf{T}_{ce,\,i}^{\text{demo}} \right\}_{i=1}^T$ is defined in the camera frame of reference, i.e., frame $\{e\}$ in frame $\{c\}$. During task execution, both the robot base and the object(s) may be relocated to different positions relative to the demonstration setup. We aim to reproduce the demonstrated trajectory while accounting for this change.

\subsubsection{Single-Object Tasks} In these tasks, the execution of the task depends only on a single manipulated object such as the ``closing door'' task in Fig.~\ref{fig:intro}. Let ${\mathbf{T}}_{co}^{\text{demo}} $ be the pose of the manipulated object $\{o\}$ estimated in the camera frame $\{c\}$ according to the human demonstration, which is the pose of the object before the human hand moves it. Let ${\mathbf{T}}_{co}^{\text{exe}}$ denote the pose of this object in the camera frame during the execution of the task. The execution trajectory $\mathcal{T}_{ce}^{\text{exe}} = \left\{ \mathbf{T}_{ce,\,i}^{\text{exe}} \right\}_{i=1}^T$ is obtained by transforming the demonstration trajectory using the object poses:
\begin{align} \label{eq:traj_transform}
    \mathcal{T}_{ce}^{\text{exe}} &= \biggl\{ \underbrace{\mathbf{T}_{co}^{\text{exe}} \cdot \left( \mathbf{T}_{co}^{\text{demo}} \right)^{-1} }_{\substack{\text{Object Pose} \\ \text{Transformation}\, \Delta \mathbf{T}}}\, \cdot \, \mathbf{T}_{ce,\,i}^{\text{demo}} \biggr\}_{i=1}^{T},
\end{align}
where $\Delta \mathbf{T} =  \mathbf{T}_{co}^{\text{exe}} \cdot \mathbf{T}_{oc}^{\text{demo}}$ is the pose change of the object between the execution of the task and the demonstration of the task. We use BundleSDF~\cite{Wen2023-bundleSDF} to compute the object pose transformation, denoted as $\Delta \mathbf{T}$. Fig.~\ref{fig:traj-in-obj-frame} illustrates how the robot gripper trajectory is transformed into a new task space by estimating this object pose transformation.

Specifically, given an RGB-D video and an object mask for the first video frame, BundleSDF tracks the object pose across subsequent frames while simultaneously reconstructing the 3D object shape based on the combined camera and object motion. In our setup, we leverage only its pose tracking capability. To construct the RGB-D video sequence for pose tracking, we select $\mathbb{M}$ frames from the human demonstration and store $\mathbb{N}$ frames from the real-time robot feed, yielding a total of $\mathbb{M}+\mathbb{N}$ frames. We replace XMem~\cite{cheng2022xmem} with SAM2~\cite{ravi2024sam2} within BundleSDF for mask generation, enabling direct use of the VIE-Object module. Preprocessed masks from the $\mathbb{M}$ demonstration frames are reused, while VIE-Object generates masks for the $\mathbb{N}$ real-time frames using the same text prompt as the demonstration.

We observe that increasing the number of frames (that is, extending the temporal context) improves the accuracy of the BundleSDF pose estimation, as it provides richer temporal information. Larger objects generally require shorter context windows, whereas smaller objects benefit from longer ones. For our experiments, we set $\mathbb{M}=10$ and $\mathbb{N}=5$. In Fig.~\ref{fig:traj-in-obj-frame}, we show for $\mathbb{M}=1$ and $\mathbb{N}=1$.
 
\subsubsection{Dual-Object Tasks}\label{dualobj-transform} In these tasks, execution depends on both the manipulated object denoted by $o_1$ and a secondary object denoted by $o_2$ in the scene such as pick-and-place tasks. The poses of the two objects can be different from the human demonstration. Therefore, we run BundleSDF~\cite{Wen2023-bundleSDF} twice to estimate the object pose transformations of $o_1$ and $o_2$. We then obtain the individual execution trajectories $\mathcal{T}_{ce_{1}}^{\text{exe}}$ and $\mathcal{T}_{ce_{2}}^{\text{exe}}$ accounting for the pose transformations of $o_1$ and $o_2$, respectively. Similar to Eq.~\eqref{eq:traj_transform}, we have
\begin{align}
\mathcal{T}_{ce_{1}}^{\text{exe}} &= \left\{ \mathbf{T}_{co_1}^{\text{exe}} \cdot \left( \mathbf{T}_{co_1}^{\text{demo}} \right)^{-1} \cdot  \mathbf{T}_{ce,\,i}^{\text{demo}} \right\}_{i=1}^{T},  \\[0.5em]
    \mathcal{T}_{ce_{2}}^{\text{exe}} &= \left\{ \mathbf{T}_{co_2}^{\text{exe}} \cdot \left( \mathbf{T}_{co_2}^{\text{demo}} \right)^{-1} \cdot  \mathbf{T}_{ce,\,i}^{\text{demo}} \right\}_{i=1}^{T}.
\end{align}
To generate a single, unified execution trajectory $\mathcal{T}_{ce}^{\text{exe}}$ for dual-object tasks, we adopt the trajectory interpolation utilized in DITTO~\cite{heppert2024ditto}. We interpolate the individual execution trajectories $\mathcal{T}_{ce_{1}}^\text{exe}$ and $\mathcal{T}_{ce_{2}}^\text{exe}$ using spherical linear interpolation (SLERP)~\cite{Shoemake1985AnimatingRW} for rotations and linear interpolation for translations. The final execution trajectory is computed as:
\begin{align}
    \mathcal{T}_{ce}^{\text{exe}} &=  \alpha(t) \cdot \mathcal{T}_{ce_{1}}^{\text{exe}} \oplus (1 - \alpha(t)) \cdot \mathcal{T}_{ce_{2}}^{\text{exe}},  \\
    \alpha(t) &= \mathcal{N}(t \mid 0, \sigma(T)) \in \mathbb{R},
\end{align}
where $\oplus$ denotes the interpolation operation, $\alpha(t) \in \mathbb{R}$ is the Gaussian weight, and $\sigma$ is a hyperparameter controlling the steepness of the mixing curve. We set $\sigma = \frac{T}{4}$ in each task for a smooth interpolation.

For both single-object and dual-object tasks, after obtaining the end-effector trajectory $\mathcal{T}_{ce}^{\text{exe}}$ in the camera frame, we then transform the trajectory into the robot base frame $\{ b \}$ using the camera pose $\mathbf{T}_{bc}^{\text{exe}}$:
\begin{align}
    \mathcal{T}_{be}^{\text{exe}} = \biggl\{ \mathbf{T}_{bc}^{\text{exe}} \cdot \mathbf{T}_{ce,i}^{\text{exe}} \biggr\}_{i=1}^{T}. \label{eq:base_traj}
\end{align}
The above trajectory serves as the reference trajectory that the end-effector of the robot needs to follow in order to complete the task. In addition, the first pose of the transferred trajectory, i.e., $\mathbf{T}_{be,1}^{\text{exe}}$ serves as the grasp pose for the manipulated object.

\subsection{Trajectory Tracking Optimization}\label{sub-sec:traj-tracking}
After transforming the end-effector trajectory into a new execution space, our goal is to ensure that the robot can follow this trajectory accurately. First, we optimize the robot base pose so that, upon reaching the optimized base pose, the execution trajectory falls approximately within the reachable space of the manipulator. Then we optimize a kinematically feasible trajectory in the joint space of the robot for the manipulator to accurately mimic the human demonstration.

\textbf{Trajectory Refinement}. 
We refine the execution trajectory $\mathcal{T}_{be}^\text{exe}$ in Eq.~\eqref{eq:base_traj} by removing outliers and excessively close points using Euclidean distance-based filtering, where the outliers may come from incorrect human hand pose estimation. This refined trajectory is then treated as the reference trajectory $\mathcal{T}^{\text{ref}} = \left\{\mathbf{T}_{be}^\text{exe}\right\}_{i=1}^{T}$ for subsequent stages.

\begin{figure}
    \centering
    \includegraphics[width=\linewidth]{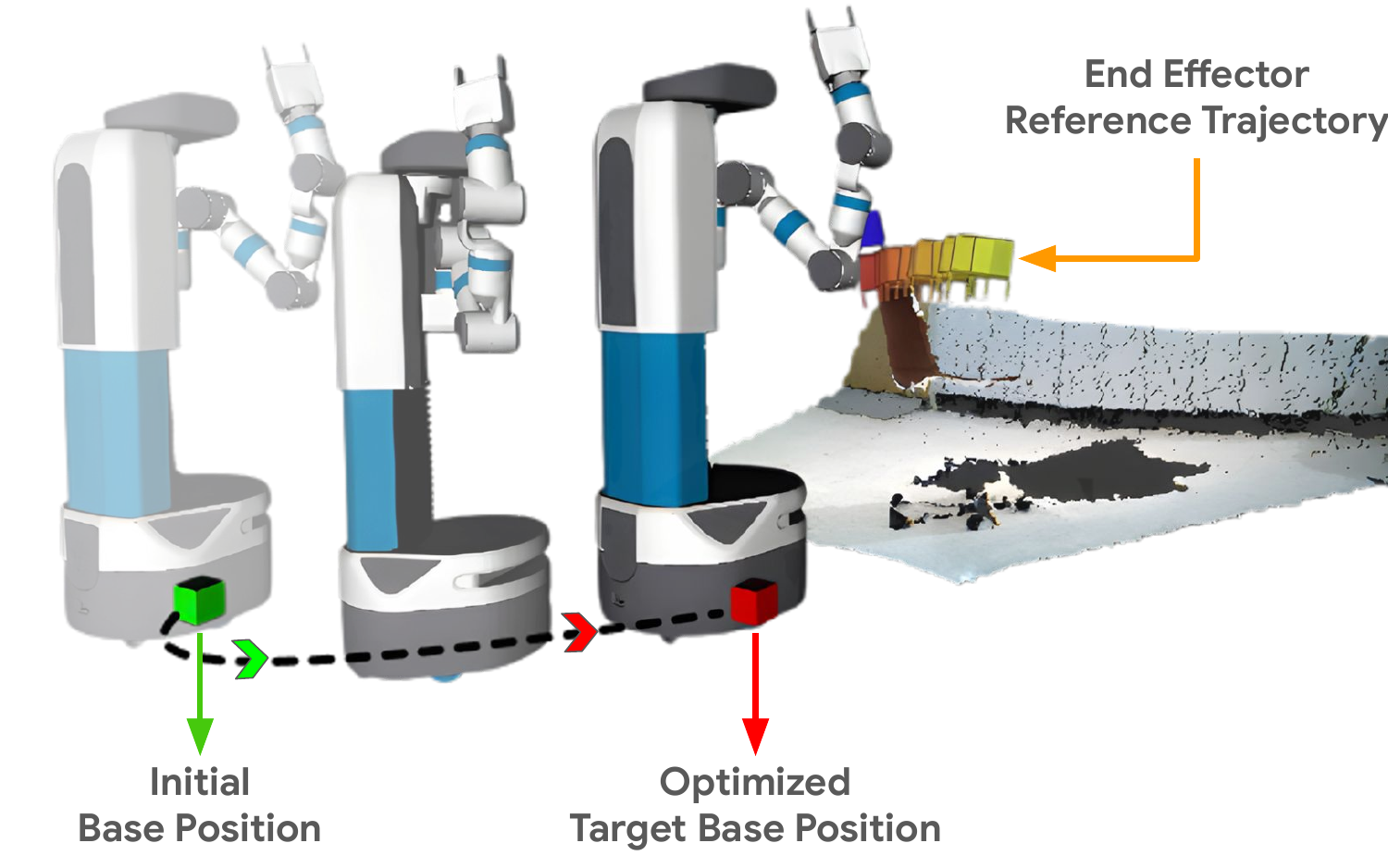}
    \vspace{-5mm}
    \caption{The figure illustrates the movement of the robot from its initial position to the optimized target position for aligning the end effector with the sampled trajectory poses}
    \label{fig:base-opt}
\end{figure}

\textbf{Stage I: Mobile base pose optimization}. 
We begin by uniformly sampling a subset of $N < T$ poses $\mathcal{T}_{s} = \{ \mathbf{T}_{be,\,i}^\text{exe}\}_{i=1}^{N}$ along the refined trajectory $\mathcal{T}^{\text{ref}}$. The sampling helps to optimize faster using fewer poses but also without losing the characteristics of the reference trajectory. The objective is to determine an optimal base pose that minimizes the cost of aligning the end-effector of the manipulator with the poses in $\mathcal{T}_{s}$. Fig.~\ref{fig:base-opt} provides a visual representation of the mobile manipulator system moving to the optimized base location. To solve this problem, first the task space for the mobile base is defined. Using the depth image from the onboard camera, we compute the 3D point cloud $\mathbf{P}_{c}$ of the scene by back-projecting the depth data using the camera intrinsics. The point cloud $\mathbf{P}_{c}$ is then transformed into the mobile base frame of reference by
\begin{align}
    \mathbf{P}_{b} &= \biggl\{ \mathbf{R}_{bc}^{\text{exe}} \cdot \mathbf{p}_i +  \mathbf{t}_{bc}^{\text{exe}} \mid \forall \, {\mathbf{p}_i} \in {\mathbf{P}_{c}} \biggr\}.
\end{align}

Next, the minimum and maximum values along the respective axes of the base frame are extracted to serve as translational constraints of the task space: 
\begin{align}
    \mathbf{x} &= [x, y, \theta]  \in \mathbb{R}^{3}, \\
    \mathbf{x}_{\min} &= [0, \hspace{0.5mm} \min_{\mathbf{p} \in {\mathbf{P}}_{b}} p_y, \hspace{0.5mm} -\pi ], \\
    \mathbf{x}_{\max} &=[ \max_{\mathbf{p} \in {\mathbf{P}}_{b}} p_x, \hspace{0.5mm} \max_{\mathbf{p} \in {\mathbf{P}}_{b}} p_y, \hspace{0.5mm} \pi ],
\end{align}
where $\mathbf{p} = [p_x, p_y, p_z]^\top$ denotes a 3D point, $(x, y)$ indicates the base position on the $x,y$ plane, and $\theta$ is the orientation of the robot base. We hope to find the optimal base configuration $\mathbf{x}$ with the lower bounds $\mathbf{x}_{\min}$ and the upper bounds $\mathbf{x}_{\max}$. Note that $\mathbf{x}$ is defined with respect to the initial base frame $\{ b \}$. Let us denote the new frame specified by $\mathbf{x}$ as frame $\{x\}$. We can convert $\mathbf{x}$ into a homogeneous transformation in the initial base frame as: 
\begin{align} \label{eq:new_base}
    \nonumber \mathbf{T}_{bx} &= (\mathbf{R}_{bx}, \mathbf{t}_{bx}) \in \mathbb{SE}(3) \\
    &= 
    \begin{bmatrix}
        \mathbf{R}_z(\theta) & [x, y, 0]^\top \\
        \mathbf{0}^\top & 1
    \end{bmatrix},
\end{align}
where $\mathbf{R}_z(\theta)$ indicates the rotation matrix around $z$-axis by $\theta$. In addition to optimizing the relative base pose change $\mathbf{x}$, we also need to optimize the manipulator configurations $\mathcal{Q} = (\mathbf{q}_1, \ldots, \mathbf{q}_N)$, where $\mathbf{q}_i \in \mathbb{R}^n, i=1,\ldots, N$ is the joint configuration corresponding to the $i$th gripper pose $\mathbf{T}_{be,\,i}^\text{exe}$ in $\mathcal{T}_{s}$, and $n$ is the degree of freedom of the manipulator. 


In order to define the objective function to optimize the base configuration $\mathbf{x}$, our idea is to check if the robot can reach these gripper poses in $\mathcal{T}_{s}$ after moving the base to $\mathbf{x}$. First, we transform the $i$th gripper pose $\mathbf{T}_{be,\,i}^\text{exe}$ to the new base frame $\{ x \}$ by
\begin{equation}
    \mathbf{T}_{xe,\,i}^\text{exe} = \mathbf{T}_{bx}^{-1} \cdot \mathbf{T}_{be,\,i}^{\text{exe}},
\end{equation}
where $\mathbf{T}_{bx}$ is given by Eq.~\eqref{eq:new_base}. Second, using forward kinematics, we can compute the gripper pose in the new base according to the joint configuration $\mathbf{q}_i$ as $\mathbf{T}_{xe}(\mathbf{q}_i)$. We would like to define a cost function $c_\text{goal}(\mathbf{T}_{xe}(\mathbf{q}_i), \mathbf{T}_{xe,\,i}^{\text{exe}})$ to minimize the distance between the two $\mathbb{SE}(3)$ transformation matrices. We utilize the goal-reaching cost in \cite{xiang2024-gto}. Let $\mathcal{E} = \{ \mathbf{p}_j \}_{j=1}^m$ be a set of $m$ 3D points sampled on the 3D mesh of the end-effector. The cost function is defined as
\begin{align} \label{eq:cost_goal}
    & c_\text{goal}(\mathbf{T}_{xe}(\mathbf{q}_i), \mathbf{T}_{xe,i}^{\text{exe}}) \nonumber \\ &= \sum_{j=1}^m\| (\mathbf{R}_{xe}(\mathbf{q}_i) \,\mathbf{p}_j + \mathbf{t}_{xe}(\mathbf{q}_i)
    - (\mathbf{R}_{xe}^{\text{exe}} \,\mathbf{p}_j + \mathbf{t}_{xe}^{\text{exe}})  \|^2,
\end{align}
which transforms the 3D points on the end effector by the two transformations and then computes the sum of the pairwise distances between the transformed points.




Finally, we formulate and solve the following constrained optimization problem to find the optimal base movement:
\begin{align} 
    \arg \min_{\mathbf{x}, \mathcal{Q}} & \Big( \lambda_{\text{effort}} \|\mathbf{x}\|^2 + \lambda_{\text{goal}} \sum_{i=1}^N c_{\text{goal}}(\mathbf{T}_{xe}(\mathbf{q}_i), \mathbf{T}_{xe,i}^{\text{exe}}) \Big) \nonumber \\
    \text{s.t.,}  &\quad -\mathbf{x}_{\text{min}} \leq \mathbf{x} \leq \mathbf{x}_{\text{max}} \nonumber \\
    &  \mathbf{q}_l \leq  \mathbf{q}_i \leq  \mathbf{q}_u,  i = 1, \ldots, N,  \label{eq:base_opt} 
\end{align}
where $\mathbf{q}_l$ and $\mathbf{q}_u$ are lower bounds and upper bounds of the joints, and $\lambda_{\text{effort}}$ and $\lambda_{\text{goal}}$ are two hyper-parameters to balance the costs. After obtaining the optimal solution $\mathbf{x^*}$ that defines a new frame $\{ x^* \}$, the mobile base is moved to the new optimized pose $\mathbf{T}_{bx^\mathbf{*}}$ according to Eq.~\eqref{eq:new_base} using a PD controller designed for non-holonomic robots.

\textbf{Stage II: Joint trajectory optimization}. Once the robot reaches the optimized base pose, we transform the reference trajectory $\mathcal{T}^{\text{ref}}$ into the new base frame of reference $\{ x^* \}$ by
\begin{equation}
     \mathcal{T}^{\text{ref}*} = \biggl\{
    \mathbf{T}_{x^*e,\,i}^\text{exe} =     
     \mathbf{T}_{bx^{*}}^{-1} \cdot \mathbf{T}_{be,i}^{\text{exe}} \biggr\}_{i=1}^{T}.
\end{equation}

Our goal now is to find a collision-free trajectory of $T$ joint configurations of the manipulator $\mathcal{Q} = (\mathbf{q}_1, \ldots, \mathbf{q}_T)$ and $T$ joint velocities $\dot{\mathcal{Q}} = (\dot{\mathbf{q}}_1, \ldots, \dot{\mathbf{q}}_T)$, where $\mathbf{q}_i, \dot{\mathbf{q}}_i \in $ $ \mathbb{R}^n$ such that the end-effector poses from forward kinematics $\left\{\mathbf{T}(\mathbf{q}_{i})\right\}_{i=1}^{T}$  align closely with $\mathcal{T}^{\text{ref}*}$. To this extent, we adopt the optimization formulation and the constraints of \cite{xiang2024-gto}, which includes a collision-avoidance term $c_{\text{collision}}(\mathbf{q}_i)$ using a point-cloud representation of the robot and the task space.


While \cite{xiang2024-gto} includes a separate penalization term for the standoff pose of grasping, we integrate the standoff pose $\mathbf{T}^{\text{standoff}}$  into the reference trajectory $\mathcal{T}^{\text{ref}*}$, eliminating the need for two separate goal-reaching terms. Since the first transformation in the reference trajectory is for grasping the target object, we compute the standoff pose as 
\begin{equation} \label{eq:standoff}
     \mathbf{T}^{\text{standoff}} = \mathbf{T}_{x^*e, 1}^{\text{exe}} \cdot \mathbf{T}_{\delta},
\end{equation}
where $\mathbf{T_{\delta}} \in \mathbb{SE}(3)$ represents a displacement along the forward axis of the gripper. Then we update the reference trajectory by appending the standoff pose to the beginning of the trajectory:
\begin{align}
    \mathcal{T}^{\text{ref}*} = \left\{ \mathbf{T}_{x^*e,i}^{\text{exe}} \right\}_{i=0}^{T}, \quad \text{with} \quad \mathbf{T}_{x^*e,0}^{\text{exe}} = \mathbf{T}^{\text{standoff}}.
\end{align}
Similarly, we append $\mathbf{q}_0$ and $\dot{\mathbf{q}}_0$ to $\mathcal{Q}$ and $\dot{\mathcal{Q}}$, respectively.
Consequently, the robot will first reach the standoff pose, then proceeds to the grasping pose for grasping, and subsequently follow the rest of the reference trajectory.
Afterwards, we solve the following constrained optimization problem to find the optimal joint trajectory:
\begin{align} 
    \arg \min_{\mathcal{Q}, \dot{\mathcal{Q}} } &  \sum_{i=0}^T \Big( \lambda c_\text{goal}(\mathbf{T}(\mathbf{q}_i), \mathbf{T}_{x^*e,i}^{\text{exe}}) \nonumber \\
    & + 
    \lambda_1 c_{\text{collision}}(\mathbf{q}_i) + \lambda_2  \| \dot{\mathbf{q}_i} \|^2 \Big) \label{eq:joint_opt} \\
    & \dot{\mathbf{q}}_0 = \mathbf{0}, \dot{\mathbf{q}}_T = \mathbf{0}  \\
    & \mathbf{q}_{i+1} = \mathbf{q}_i + \dot{\mathbf{q}}_i dt, i = 0, \ldots, T-1  \label{eq:dynamics}\\
    &  \mathbf{q}_l \leq  \mathbf{q}_i \leq  \mathbf{q}_u,  i = 0, \ldots, T  \label{eq:p_bound} \\
    & \dot{\mathbf{q}}_l \leq  \dot{\mathbf{q}}_i \leq  \dot{\mathbf{q}}_u,  i = 0, \ldots, T, \label{eq:v_bound}
\end{align}
where $c_{\text{goal}}$ and $c_{\text{collision}}$ are defined according to~\cite{xiang2024-gto}, and $\lambda, \lambda_1, \lambda_2$ are hyper-parameters. For the constraints: 1) $\dot{\mathbf{q}}_0 = \mathbf{0}, \dot{\mathbf{q}}_T = \mathbf{0}$ ensure that the starting velocity and the ending velocity of the robot are zero. 2) $\mathbf{q}_{i+1} = \mathbf{q}_i + \dot{\mathbf{q}}_i dt$ ensures that the robot state follows the kinematics of the robot, where $dt$ is the time interval between two time steps. 3) The last two constraints in Eqs.~\eqref{eq:p_bound} and \eqref{eq:v_bound} ensure that the joint positions and the joint velocities are within the lower bounds $(\mathbf{q}_l, \dot{\mathbf{q}}_l)$ and the upper bounds $(\mathbf{q}_u, \dot{\mathbf{q}}_u)$.

\begin{figure}
    \centering
    \includegraphics[width=\linewidth]{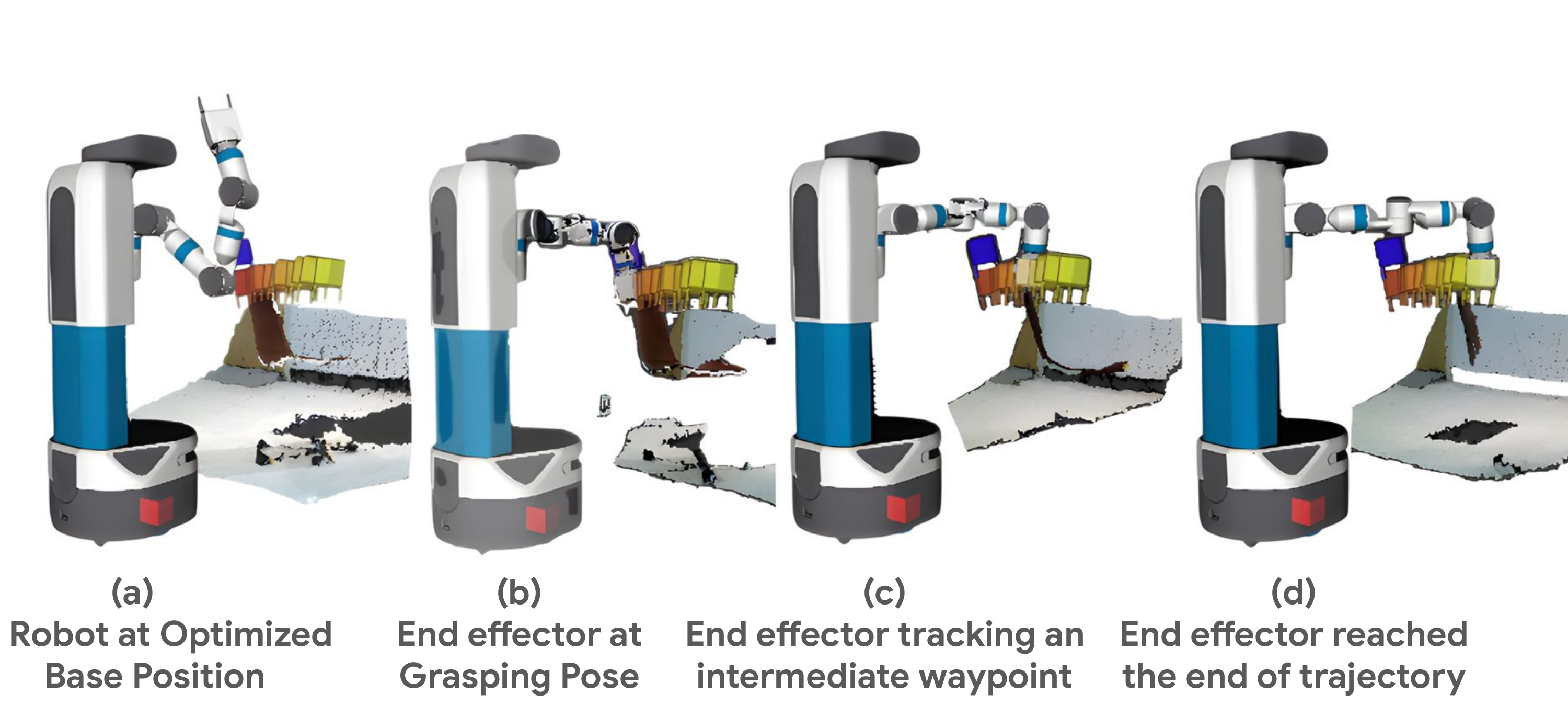}
    \vspace{-4mm}
    \caption{Illustration of the execution of an optimized trajectory. The blue gripper indicates the standoff pose for grasping.}
    \label{fig:trajopt}
    \vspace{-1mm}
\end{figure}

By solving the above optimization problem, we can find the joint trajectory that follows the reference end-effector trajectory $\mathcal{T}^{\text{ref}*}$. Consequently, the robot can perform the same task as in the human demonstration. Fig.~\ref{fig:trajopt} illustrates an example of the fetch robot following a reference end-effector trajectory, where the blue gripper indicates the standoff pose for grasping. We utilize the Interior Point OPTimizer (Ipopt)~\cite{wachter2006implementation}  with the CasADi framework~\cite{Andersson2019} to solve both optimization problems in Eqs.~\eqref{eq:base_opt} and \eqref{eq:joint_opt}. 

%% file: includes/sections/4-experiments.tex
\section{Experiments}\label{sec:experiments}

All experiments are conducted using a Fetch mobile manipulator robot, equipped with a parallel jaw gripper. To assess the performance and generalization of our method, we conducted trials in a variety of indoor environments.  The system is developed using ROS Noetic, and MoveIt~\cite{chitta2012moveit} is utilized to execute the manipulation trajectories. All perception, optimization, and control modules run online on a laptop with a NVIDIA RTX 4090 GPU, which is connected to the robot during execution.

\subsection{Tasks}\label{sub-sec:tasks}

Most of the objects in this work come from the \textsc{FewSOL}~\cite{padalunkal2023fewsol} dataset, which includes everyday items for robotic use. 16 tasks (see Table~\ref{tab: ditto-hrt1-results}) are designed to cover common household activities, from simple tasks like picking and placing objects to more difficult tasks like inserting objects. These 16 tasks include single-object tasks (e.g., moving a chair ) and dual-object tasks (e.g., putting a bread in a toaster, cleaning a plate with a brush, or hanging a cap on a hook). Each task is executed three times with different object(s) placements each time and in an environment that is significantly different from the demonstration.

\subsection{Demonstration data Collection and Processing}
We collect human demonstrations using the mechanism described in ~Sec.\ref{sub-sec:data-collection}. Each demonstration is approximately 7 to 12 seconds, collected at $\sim$ 10fps. We observed that when the human demonstrator grasps the object arbitrarily (e.g., using fingertip contacts), in-hand object manipulation could occur, and the transferred grasp may not be reliable for the parallel jaw robot gripper. To mitigate this issue, we moved our hand slightly closer to the object during grasping, allowing a more stable transferred grasp pose, while also ensuring better finger visibility for hand pose estimation.

\subsection{Manual Object Pose Verification}\label{sub-sec:pose-verify}
\begin{figure}[!h]
    \centering
    \includegraphics[width=\linewidth,trim={0 2cm 0.05cm 0.2cm},clip]{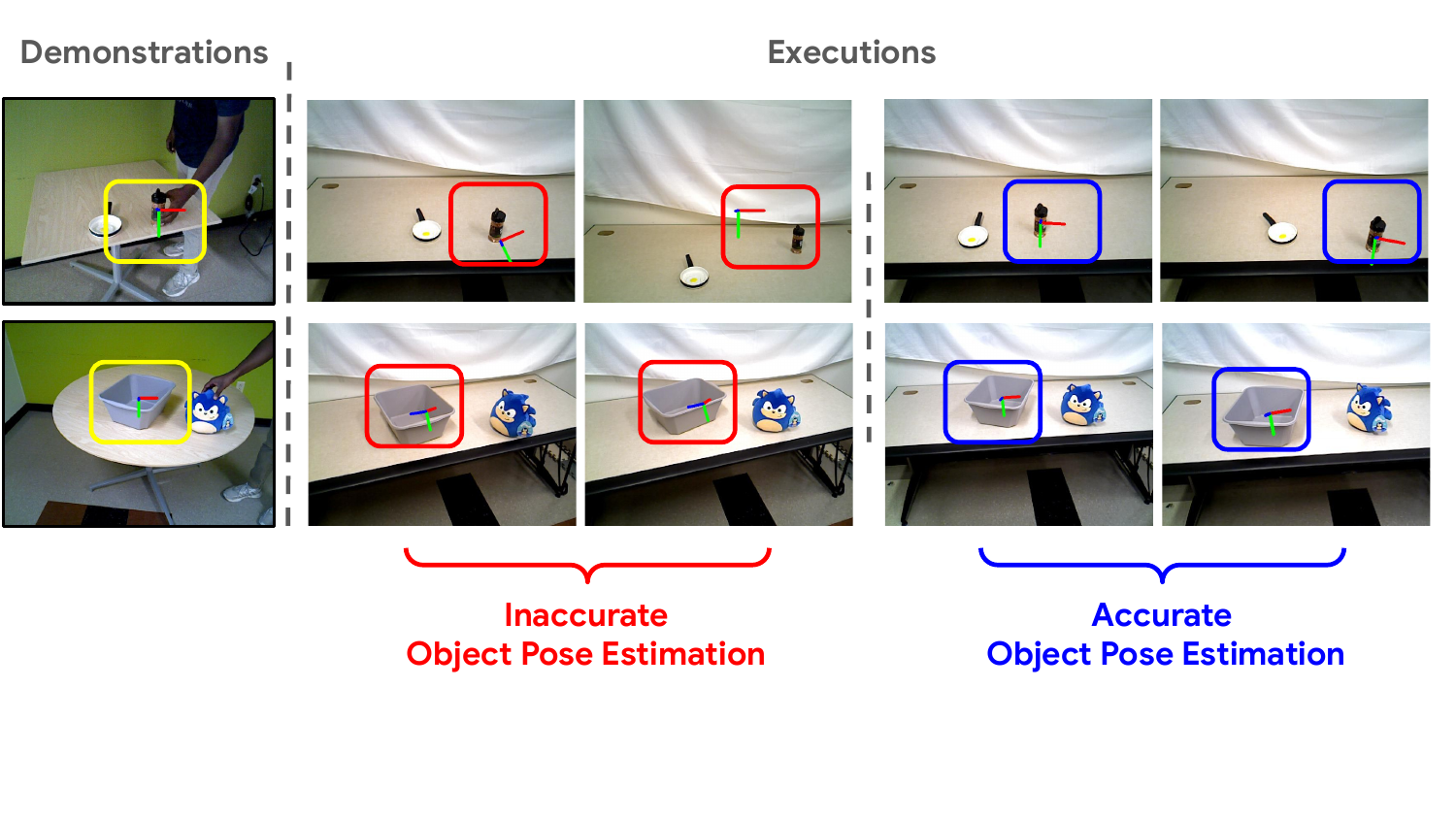}
    \vspace{-5mm}
    \caption{Example scenarios of accurate and inaccurate object pose estimations using BundleSDF~\cite{Wen2023-bundleSDF} during the execution of tasks: \textit{Put seasoning on the omelette} (top), \textit{Put toy in the bin} (bottom) compared to their demonstration scenes.}
    \label{fig:manual_pose_verify}
\end{figure}

As illustrated in Fig.~\ref{fig:traj-in-obj-frame}, we compute the object pose transformation to align the demonstration trajectory with the robot task space. However, in some cases, BundleSDF~\cite{Wen2023-bundleSDF} fails to accurately estimate the object pose, due to poor lighting, holes in depth images, and large object rotations relative to the demonstration. The object pose errors result in inaccurate trajectories, where the task execution is very likely to fail. Examples of inaccurate pose estimation and accurate pose estimation are shown in Fig.~\ref{fig:manual_pose_verify}. Therefore, during our experiments, before executing the transformed trajectory, we visually inspect the corresponding object pose axes to decide whether to proceed or not. If the object pose is visually incorrect, we rearrange the object(s) placements. We apply this manual pose verification procedure during the execution of both our method and the baseline method. We only execute the trajectory when the object pose looks correct in order to mitigate the effect of object pose estimation on the system.


\subsection{Baseline}\label{sub-sec:baseline}
To evaluate our system, we benchmark its performance against DITTO~\cite{heppert2024ditto}, which transfers the trajectory of the object poses in the human demonstration to the robot. DITTO~\cite{heppert2024ditto} does not provide a complete implementation code for real-world execution with a robot and only offers the code to extract object trajectories for execution. Hence, we implement the real-world execution based on their approach. 

For single-object tasks, first, we leverage BundleSDF~\cite{Wen2023-bundleSDF} on the human demonstration data to extract the 6-DoF pose of the manipulated object in each frame. Then we obtain the manipulated object trajectory $\mathcal{T}_{co}^{\text{demo}} = \{ \mathbf{T}_{co,i}^{\text{demo}}  \}_{i=1}^T$ and the corresponding robot grasp pose $\mathbf{T}_{ce,1}^{\text{demo}}$ from the demonstration trajectory $\mathcal{T}_{ce}^\text{demo}$, where the first gripper pose is for grasping. During execution, the object trajectory and the grasp pose undergo a pose transformation (see Sec.~\ref{sub-sec:traj-in-obj-frame}) to get the current object trajectory in the robot base $\mathcal{T}_{bo}^{\text{exe}}$, and the current transferred grasp pose $\mathbf{T}_{be,1}^{\text{exe}}$ during execution. Next, using GroundingDINO~\cite{di2024dinobot} and SAM2~\cite{ravi2024sam2}, we obtain the manipulated object mask. Applying this mask to the point cloud from the depth image $P_b$, we extract the point cloud of the manipulated object $P_{bo}$. We then run Contact-GraspNet~\cite{9561877} on $P_{bo}$ to generate a set of candidate grasps $\mathcal{G}$. From $\mathcal{G}$, we select the grasp $\mathbf{T}_{g^*}$ having the smallest distance to the transferred grasp pose $\mathbf{T}_{be,1}^{\text{exe}}$. We then compute the end-effector trajectory using the object trajectory, by applying the transformation between the  current object pose $\mathbf{T}_{bo,1}^{\text{exe}}$ and the selected grasp $\mathbf{T}_{g^*}$.

Specifically, we compute
\begin{align}
    &\mathbf{T}_{og^{*}} =  {(\mathbf{T}_{bo,1}^{\text{exe}}})^{-1} \cdot \mathbf{T}_{g^*}, \\
    &\mathbf{T}^{\text{standoff}} = \mathbf{T}_{g^{*}} \cdot \mathbf{T}_{\delta}, \\
    &\mathcal{T}^{\text{ref}}  = \mathcal{T}_{be}^{\text{exe}} = \biggl\{ \mathbf{T}^{\text{standoff}},  \mathbf{T}_{bo,\,i}^{\text{exe}} \, \cdot  \mathbf{T}_{og^{*}} \biggr
    \}_{i=1}^{T},
\end{align}
where $\mathcal{T}^{\text{ref}}$ is the end-effector trajectory before the robot base optimization. 

For dual object tasks, we combine the manipulated and secondary object trajectories as described in Sec.~\ref{dualobj-transform}. The resulting execution trajectory $\mathcal{T}_{\text{ref}}$ is used for base optimization. Once the robot reaches the optimized base position, we compute the final trajectory in the new robot base as $\mathcal{T}_{\text{ref}}^{\textbf{*}}$. At this point, following DITTO~\cite{heppert2024ditto}, we compute Inverse Kinematics for each pose in the trajectory using the KDL Kinematics solver in MoveIt~\cite{chitta2012moveit} and execute all the valid target poses. Since the baseline does not include any obstacle avoidance mechanism, we integrate two safety features during execution. First, we constrain the optimized base solution to prevent the robot from moving beyond the target object, avoiding potential collision. Second, we leverage the force sensor installed in our manipulator arm, to detect excessive contact force during a hard contact/collision with the environment. In such cases, the execution is stopped, and task is considered to be a failure.

\subsection{Evaluation Metrics}\label{sub-sec:metrics}

To quantitatively evaluate and compare the performance of our proposed method against the baseline and the accuracy of the optimization framework, we utilize the following metrics.
\begin{itemize}
    \item \textbf{Grasp success}: This is a binary metric that determines whether the end-effector securely grasps the target object. A grasp is treated as successful if the object is lifted from its surface without slipping or falling.
    
   \item \textbf{Task Completion}: This metric is also binary, which evaluates the success of the entire manipulation task. A task is considered successful if the robot grasps the object and follows the trajectory to achieve the semantic goal of the task, i.e., its intended purpose.
   
    \item \textbf{Tracking Error}: This metric quantifies the accuracy of following a reference trajectory by calculating the average error between the executed and desired end-effector poses. Given an executed trajectory $\mathcal{T}_1 = \{(\mathbf{R}_i^{1}, \mathbf{t}_i^{1})\}_{i=1}^{T}$ and a reference trajectory $\mathcal{T}_2 = \{(\mathbf{R}_i^{2}, \mathbf{t}_i^{2})\}_{i=1}^{T}$, the tracking error is decomposed into two components:

    \begin{itemize}
        \item \textit{Translation Error} ($\mathcal{E}_{\text{trans}}$): The root mean square error (RMSE) between the corresponding translation vectors:
        \[
        \mathcal{E}_{\text{trans}} = \sqrt{\frac{1}{T} \sum_{i=1}^{T} \left\| \mathbf{t}_i^{1} - \mathbf{t}_i^{2} \right\|_2^2}.
        \]
        \item \textit{Rotation Error} ($\mathcal{E}_{\text{rot}}$): The RMSE of the geodesic distance (angle) between the corresponding orientations:
       \[
\mathcal{E}_{\mathrm{rot}} =
\sqrt{\frac{1}{T} \sum_{i=1}^{T} 
\Bigl( \arccos \frac{\mathrm{tr}(\mathbf{R}_i^{1} (\mathbf{R}_i^{2})^\top) - 1}{2} \Bigr)^2 }.
\]

    \end{itemize}
    
Grasp success and task completion are decided manually during the experiments.
\end{itemize}

\subsection{Optimization}
The hyper parameters for the optimization problems in Sec.~\ref{sub-sec:traj-tracking} are set as follows. In Eq.~\eqref{eq:base_opt}. $\lambda_{\text{effort}}=0.01$, $\lambda_{\text{goal}}=1$. In Eq.~\eqref{eq:joint_opt}, $\lambda=150$, $\lambda_{1}=0.02$, $\lambda_{2}=0.01$. The standoff pose $\mathbf{T}^{\text{standoff}}$ in Eq.~\eqref{eq:standoff} is set to be 20cm away from the grasping pose, along the gripper forward axis. We run both the mobile base pose optimization and the joint trajectory optimization for 100 iterations each, with a tolerance of $1e^{-15}$. We then use TOPP-RA~\cite{article} to time-parameterize the optimized joint trajectory and execute a smooth and kinematically feasible motion after grasping the object.

\subsection{Results}\label{sub-sec:results}

 \begin{figure*}[h]
     \centering
     \includegraphics[width=0.9\linewidth]{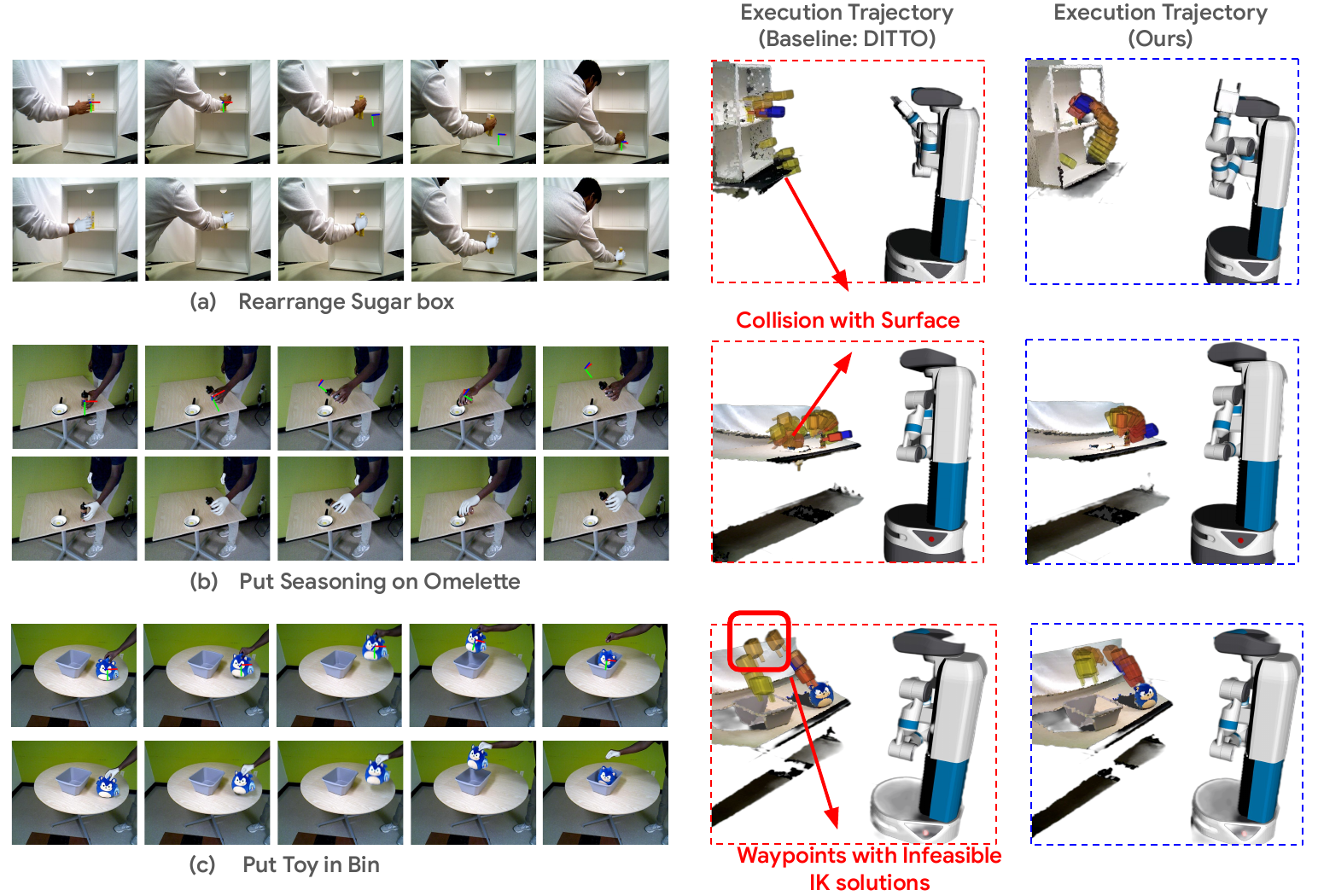}
     \caption{Execution trajectories of three tasks:  object-based trajectories for the baseline DITTO~\cite{heppert2024ditto} vs. hand-based trajectories in our system. Our method produces smoother, more reliable motions, while the baseline often shows infeasible pose targets.}
     \label{fig:result-traj}
 \end{figure*}

We evaluated our proposed framework, HRT1, against the baseline DITTO~\cite{heppert2024ditto} though real world experiments. the evaluation was conducted across 16 tasks, categorized into single-object and dual-object settings. For each task, we performed 3 trials and measured the performance using the \textbf{Grasp Success} and \textbf{Task Completion} metrics. The experimental results are summarized in Table \ref{tab: ditto-hrt1-results}. 

Our system demonstrated a significant improvement over baseline in both metrics. Our method achieved ~98\% (47/48) grasp success rate, compared to only 58\% (28/48) for the baseline. The performance gasp is even more pronounced in task completion. We achieved task success in roughly 92\% (44/48) of the trials, and the baseline managed to succeed in only 21\% (10/48) of the trials. 

\begin{table}[ht]
\centering
\resizebox{\columnwidth}{!}{%
\begin{tabular}{ccccc}
\hline
\multicolumn{1}{l|}{\multirow{2}{*}{Skill}} & \multicolumn{2}{c|}{Grasp success} & \multicolumn{2}{c}{Task completion} \\ \cline{2-3} \cline{4-5} 
\multicolumn{1}{l|}{} & \multicolumn{1}{c|}{DITTO~\cite{heppert2024ditto}} & \multicolumn{1}{c|}{Ours} & \multicolumn{1}{c|}{DITTO~\cite{heppert2024ditto}} & \multicolumn{1}{c}{Ours} \\ \hline
\multicolumn{5}{c}{\textbf{Single object}} \\ \hline
\multicolumn{1}{l|}{Move the chair} & \multicolumn{1}{c|}{{\color[HTML]{32CB00} \textbf{3}}} & \multicolumn{1}{c|}{{\color[HTML]{32CB00} \textbf{3}}} & \multicolumn{1}{c|}{{\color[HTML]{FE0000} \textbf{0}}} & \multicolumn{1}{c}{{\color[HTML]{32CB00} \textbf{3}}} \\ \hline
\multicolumn{1}{l|}{Close fire extinguisher door} & \multicolumn{1}{c|}{{\color[HTML]{FE0000} \textbf{0}}} & \multicolumn{1}{c|}{{\color[HTML]{32CB00} \textbf{3}}} & \multicolumn{1}{c|}{{\color[HTML]{FE0000} \textbf{0}}} & \multicolumn{1}{c}{{\color[HTML]{32CB00} \textbf{3}}} \\ \hline
\multicolumn{5}{c}{\textbf{Dual object}} \\ \hline
\multicolumn{1}{l|}{Put toy in the bin} & \multicolumn{1}{c|}{{\color[HTML]{FFAA00} \textbf{2}}} & \multicolumn{1}{c|}{{\color[HTML]{32CB00} \textbf{3}}} & \multicolumn{1}{c|}{{\color[HTML]{FE0000} \textbf{1}}} & \multicolumn{1}{c}{{\color[HTML]{32CB00} \textbf{3}}} \\ \hline
\multicolumn{1}{l|}{Put bread in the toaster} & \multicolumn{1}{c|}{{\color[HTML]{FE0000} \textbf{1}}} & \multicolumn{1}{c|}{{\color[HTML]{32CB00} \textbf{3}}} & \multicolumn{1}{c|}{{\color[HTML]{FE0000} \textbf{1}}} & \multicolumn{1}{c}{{\color[HTML]{32CB00} \textbf{3}}} \\ \hline
\multicolumn{1}{l|}{Put seasoning on the omelette} & \multicolumn{1}{c|}{{\color[HTML]{32CB00} \textbf{3}}} & \multicolumn{1}{c|}{{\color[HTML]{32CB00} \textbf{3}}} & \multicolumn{1}{c|}{{\color[HTML]{FFAA00} \textbf{2}}} & \multicolumn{1}{c}{{\color[HTML]{32CB00} \textbf{3}}} \\ \hline
\multicolumn{1}{l|}{Put Lays on the red plate} & \multicolumn{1}{c|}{{\color[HTML]{FFAA00} \textbf{2}}} & \multicolumn{1}{c|}{{\color[HTML]{FFAA00} \textbf{2}}} & \multicolumn{1}{c|}{{\color[HTML]{FE0000} \textbf{1}}} & \multicolumn{1}{c}{{\color[HTML]{FFAA00} \textbf{2}}} \\ \hline
\multicolumn{1}{l|}{Clean plate with brush} & \multicolumn{1}{c|}{{\color[HTML]{FE0000} \textbf{1}}} & \multicolumn{1}{c|}{{\color[HTML]{32CB00} \textbf{3}}} & \multicolumn{1}{c|}{{\color[HTML]{FE0000} \textbf{0}}} & \multicolumn{1}{c}{{\color[HTML]{32CB00} \textbf{3}}} \\ \hline
\multicolumn{1}{l|}{Clean plate with tissue} & \multicolumn{1}{c|}{{\color[HTML]{FE0000} \textbf{0}}} & \multicolumn{1}{c|}{{\color[HTML]{32CB00} \textbf{3}}} & \multicolumn{1}{c|}{{\color[HTML]{FE0000} \textbf{0}}} & \multicolumn{1}{c}{{\color[HTML]{32CB00} \textbf{3}}} \\ \hline
\multicolumn{1}{l|}{Clean plate with kitchen towel} & \multicolumn{1}{c|}{{\color[HTML]{FFAA00} \textbf{2}}} & \multicolumn{1}{c|}{{\color[HTML]{32CB00} \textbf{3}}} & \multicolumn{1}{c|}{{\color[HTML]{FE0000} \textbf{1}}} & \multicolumn{1}{c}{{\color[HTML]{32CB00} \textbf{3}}} \\ \hline
\multicolumn{1}{l|}{Remove cap from wall hook} & \multicolumn{1}{c|}{{\color[HTML]{32CB00} \textbf{3}}} & \multicolumn{1}{c|}{{\color[HTML]{32CB00} \textbf{3}}} & \multicolumn{1}{c|}{{\color[HTML]{FE0000} \textbf{1}}} & \multicolumn{1}{c}{{\color[HTML]{32CB00} \textbf{3}}} \\ \hline
\multicolumn{1}{l|}{Hang cap onto wall hook} & \multicolumn{1}{c|}{{\color[HTML]{FE0000} \textbf{0}}} & \multicolumn{1}{c|}{{\color[HTML]{32CB00} \textbf{3}}} & \multicolumn{1}{c|}{{\color[HTML]{FE0000} \textbf{0}}} & \multicolumn{1}{c}{{\color[HTML]{FFAA00} \textbf{2}}} \\ \hline
\multicolumn{1}{l|}{Take out sugar box from shelf} & \multicolumn{1}{c|}{{\color[HTML]{FE0000} \textbf{1}}} & \multicolumn{1}{c|}{{\color[HTML]{32CB00} \textbf{3}}} & \multicolumn{1}{c|}{{\color[HTML]{FE0000} \textbf{0}}} & \multicolumn{1}{c}{{\color[HTML]{32CB00} \textbf{3}}} \\ \hline
\multicolumn{1}{l|}{Rearrange sugar box in the shelf} & \multicolumn{1}{c|}{{\color[HTML]{FFAA00} \textbf{2}}} & \multicolumn{1}{c|}{{\color[HTML]{32CB00} \textbf{3}}} & \multicolumn{1}{c|}{{\color[HTML]{FE0000} \textbf{0}}} & \multicolumn{1}{c}{{\color[HTML]{FFAA00} \textbf{2}}} \\ \hline
\multicolumn{1}{l|}{Place bottle in the shelf} & \multicolumn{1}{c|}{{\color[HTML]{32CB00} \textbf{3}}} & \multicolumn{1}{c|}{{\color[HTML]{32CB00} \textbf{3}}} & \multicolumn{1}{c|}{{\color[HTML]{FE0000} \textbf{0}}} & \multicolumn{1}{c}{{\color[HTML]{32CB00} \textbf{3}}} \\ \hline
\multicolumn{1}{l|}{Close jar with a lid} & \multicolumn{1}{c|}{{\color[HTML]{FFAA00} \textbf{2}}} & \multicolumn{1}{c|}{{\color[HTML]{32CB00} \textbf{3}}} & \multicolumn{1}{c|}{{\color[HTML]{FE0000} \textbf{0}}} & \multicolumn{1}{c}{{\color[HTML]{FFAA00} \textbf{2}}} \\ \hline
\multicolumn{1}{l|}{Displace cracker box} & \multicolumn{1}{c|}{{\color[HTML]{32CB00} \textbf{3}}} & \multicolumn{1}{c|}{{\color[HTML]{32CB00} \textbf{3}}} & \multicolumn{1}{c|}{{\color[HTML]{32CB00} \textbf{3}}} & \multicolumn{1}{c}{{\color[HTML]{32CB00} \textbf{3}}} \\ \hline
\hline
\multicolumn{1}{l|}{\textbf{Total}} & \multicolumn{1}{c|}{\textbf{28/48}} & \multicolumn{1}{c|}{\textbf{47/48}} & \multicolumn{1}{c|}{\textbf{10/48}} & \multicolumn{1}{c}{\textbf{44/48}} \\ \hline
\end{tabular}%
}
\caption{Comparison of DITTO~\cite{heppert2024ditto} and HRT1 on grasp and task success. Task completion is verified manually and it is only considered till the end of trajectory.}
\label{tab: ditto-hrt1-results}
\vspace{-4mm}
\end{table}

\textbf{Robustness of Demonstration Trajectories}. The primary reason for the observed performance gap in task execution lies in the procedure used to obtain demonstration trajectories. Our approach leverages hand poses to get the trajectory of transferred grasps. Since the human hand is clearly visible in most frames, hand pose predictions are stable, resulting in a valid and reliable transferred trajectory. Meanwhile, the baseline DITTO~\cite{heppert2024ditto} is based on object poses. When a human performs the task naturally, the target objects are often occluded by the hand. Moreover, object pose tracking depends on the pose prediction of the previous frame, so rapid hand movements can cause object tracking failures as well. This significantly degrades the quality of the object pose estimation from BundleSDF~\cite{Wen2023-bundleSDF} leading to erroneous demonstration trajectories for DITTO~\cite{heppert2024ditto}. Fig.~\ref{fig:result-traj} qualitatively illustrates this issue in several tasks, showing that our hand poses remain stable while object pose tracking is erratic due to occlusions.

This flawed demonstration trajectory of the object poses transforms into an unworkable execution plan for the robot. For example, in the ``\textit{rearrange sugar box}'' task, shown in Fig.~\ref{fig:result-traj}(a), the trajectory generated by DITTO~\cite{heppert2024ditto} is jerky and inaccurate, while using our method, the execution trajectory is smooth and accurate. An even more critical failure is observed in the ``\textit{put seasoning on omelette}'' task as shown in Fig.~\ref{fig:result-traj}(b), where some target poses collide with the surface of the table making them infeasible to execute. In contrast, by tracking the consistently visible hand, our method generates smooth and valid trajectories that enables successful task executions. Although object pose tracking can be effective in tasks with minimal occlusions, such as ``\textit{put the toy in bin}'' shown in Fig.~\ref{fig:result-traj}(c), execution of the baseline failed mainly due to some waypoints in the trajectory with infeasible inverse kinematics solutions, arising from grasp pose offset, as discussed next. 

 \begin{figure}[ht]
     \centering
     \includegraphics[width=\linewidth,trim={0, 5cm, 15cm, 0.7cm},clip]{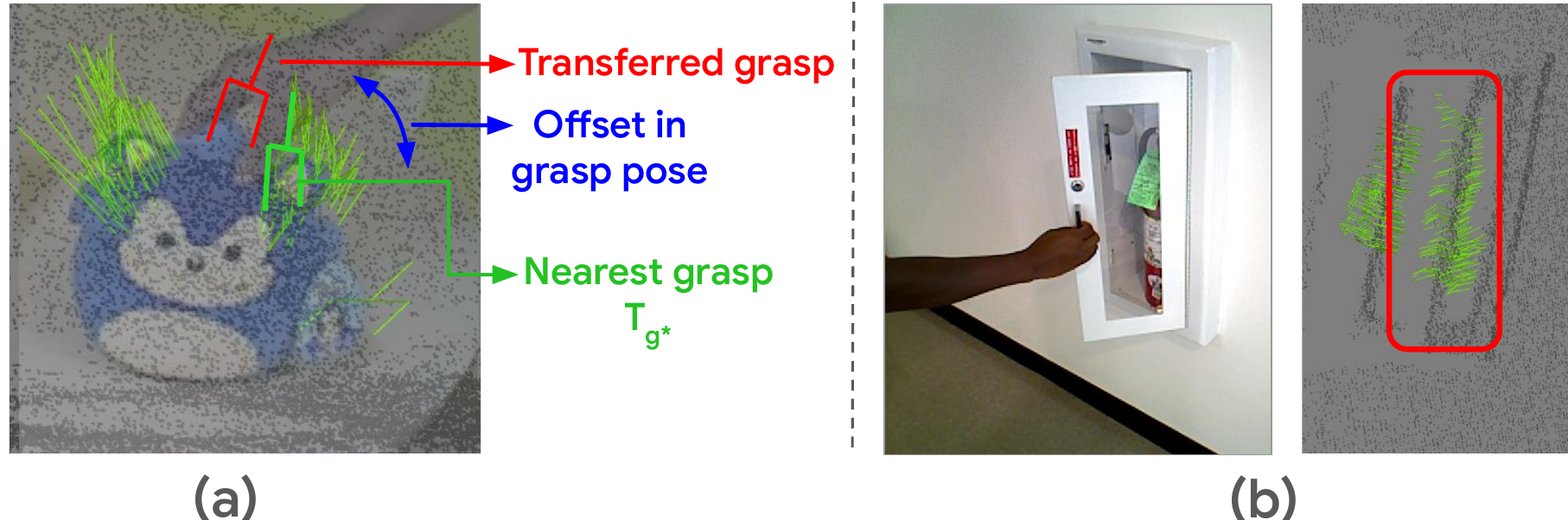}
     \vspace{-1mm}
     \caption{(a) Offset in grasping pose induced during the selection of nearest grasp during execution of baseline. (b) Human demonstration of grasping the door at the handle (left) and grasps of door generated by Contact-GraspNet~\cite{9561877} during execution (right).}
     \label{fig:offset-grasping}

     \vspace{-4mm}
 \end{figure}

\begin{table}[!h]
\resizebox{0.9\columnwidth}{!}{%
\begin{tabular}{ccc}
\hline
\multicolumn{1}{c|}{\multirow{2}{*}{\textbf{Skill}}} & \multicolumn{2}{c}{\textbf{Trajectory pose errors}} \\ \cline{2-3} 
\multicolumn{1}{c|}{} & \multicolumn{1}{c|}{$\mathcal{E}_{\text{rot}}$ \textbf{{[}rad{]}}} & $\mathcal{E}_{\text{trans}}$ \textbf{{[}m{]}} \\ \hline
\multicolumn{3}{c}{\textbf{Single object}} \\ \hline
\multicolumn{1}{l|}{Move the chair} & \multicolumn{1}{c|}{0.0386} & 0.0062 \\ \hline
\multicolumn{1}{l|}{Close fire extinguisher door} & \multicolumn{1}{c|}{0.0112} &  0.0025\\ \hline
\multicolumn{3}{c}{\textbf{Dual object}} \\ \hline
\multicolumn{1}{l|}{Put toy in the bin} & \multicolumn{1}{c|}{0.3339} &  0.0103\\ \hline
\multicolumn{1}{l|}{Put bread in the toaster} & \multicolumn{1}{c|}{0.2735} & 0.0036 \\ \hline
\multicolumn{1}{l|}{Put seasoning on the omelette} & \multicolumn{1}{c|}{0.0812} & 0.0025 \\ \hline
\multicolumn{1}{l|}{Put Lays on the red plate} & \multicolumn{1}{c|}{0.1720} &  0.0059\\ \hline
\multicolumn{1}{l|}{Clean plate with brush} & \multicolumn{1}{c|}{0.047} &  0.0024\\ \hline
\multicolumn{1}{l|}{Clean plate with tissue} & \multicolumn{1}{c|}{0.0862} & 0.0032 \\ \hline
\multicolumn{1}{l|}{Clean plate with kitchen towel} & \multicolumn{1}{c|}{0.0868} & 0.0024 \\ \hline
\multicolumn{1}{l|}{Remove cap from wall hook} & \multicolumn{1}{c|}{0.0307} & 0.0029 \\ \hline
\multicolumn{1}{l|}{Hang cap onto wall hook} & \multicolumn{1}{c|}{0.1435} & 0.0031 \\ \hline
\multicolumn{1}{l|}{Take out sugar box from shelf} & \multicolumn{1}{c|}{0.3119} &  0.0022\\ \hline
\multicolumn{1}{l|}{Rearrange sugar box in the shelf} & \multicolumn{1}{c|}{0.0563} & 0.0036 \\ \hline
\multicolumn{1}{l|}{Place bottle in the shelf} & \multicolumn{1}{c|}{0.0814} &  0.0065\\ \hline
\multicolumn{1}{l|}{Close jar with a lid} & \multicolumn{1}{c|}{0.1116} & 0.0026 \\ \hline
\multicolumn{1}{l|}{Displace cracker box} & \multicolumn{1}{c|}{0.0418} &  0.0025\\ \hline \hline
\multicolumn{1}{l|}{\textbf{Average}} & \multicolumn{1}{c|}{\textbf{0.1192}} &  \textbf{0.0038}\\ \hline
\end{tabular}%
}
\caption{Metrics for trajectory tracking of 16 tasks in our experiments.}
\vspace{-5mm}
\label{tab: metrics}
\end{table}
 
\textbf{Grasp Selection and Error Propagation}. Another critical limitation of the baseline is its grasp generation and selection mechanism. DITTO~\cite{heppert2024ditto} uses Contact-GraspNet~\cite{9561877} to generate grasps from the object point cloud. While the algorithm selects $\mathbf{T}_{g^{*}}$, in the vicinity of the grasp demonstrated by humans, this can induce some significant offset in the grasp pose, especially if there are no grasps generated in the immediate vicinity of the grasp pose demonstrated by humans, as shown in Fig.~\ref{fig:offset-grasping}(a). This grasp offset propagates through the entire trajectory, pushing several poses beyond the reachable space of the robot, which results in inverse kinematics failures during execution. The ``\textit{put toy in bin}'' task clearly demonstrated this failure, as shown in Fig.~\ref{fig:result-traj}(c). This also explains many failures of the baseline in tasks requiring precise placement such as ``\textit{hang cap onto wall hook}, \textit{close jar with a lid}, \textit{place bread in toaster}''. Moreover, grasp selection highly depends on the object point cloud. Fig.~\ref{fig:offset-grasping}(b) shows, in the task of ``\textit{close fire extinguisher door}'', infeasible grasps are generated inside the frame of the door, leading to grasp failures. This is because of the lack of depth in the transparent region of the door, which results in an incomplete point cloud.

Our method avoids this problematic grasp selection process by directly using the transferred end-effector poses from the human hand. Thus, we ensure that the initial grasp and the subsequent trajectory waypoints are reliable. In summary, the superior performance of our method HRT1, is attributed to its robust trajectory generation via grasp transfer through hand poses and the use of the two-stage optimization process to track the trajectory accurately.

To quantitatively measure the performance of our trajectory tracking, we measure the rotation and translation errors between the target and the executed end-effector paths during real-world executions. As shown in Table~\ref{tab: metrics}, our two-stage optimization process maintains low tracking errors in all tasks, with an average rotational error of 0.1192 radians and an average translational error of 0.0038 meters. These results demonstrate that our method can accurately follow the human trajectory transferred. We notice that a small fraction of these errors, especially in rotation, arise due to minor odometry drift of the mobile base of the robot during task execution.

\subsection{Runtime} 

Across all experiments, the trajectory alignment step (Sec.~\ref{sub-sec:traj-in-obj-frame}), has an average runtime of $\sim$9.6s per object in the task, where $\sim$1s is spent on object detection and segmentation of 5 real-time frames using GroundingDINO~\cite{GDINO} and SAM2~\cite{ravi2024sam2} and $\sim$8.6s for relative pose estimation using 10 demonstration frames and the above 5 real-time frames in BundleSDF~\cite{Wen2023-bundleSDF}. The optimization of the mobile base pose takes $\sim$2.2s on average, while the optimization of the joint trajectory takes $\sim$2.8s.  


%% file: includes/sections/5-conclusion.tex
\section{Conclusion} \label{sec:conclusion}
In this work, we introduce HRT1, a training-free framework for one-shot human-to-robot trajectory transfer in mobile manipulation. HRT1 enables a mobile manipulation robot to imitate a human demonstration directly from an RGB-D video, without task-specific training. Our framework leverages 3D human hand pose estimation and grasp transfer to generate a robust robot end-effector trajectory for the task. Furthermore, we introduce a two-stage optimization algorithm that solves for optimal base positioning and precise end-effector trajectory tracking for task execution in the real world. We have conducted real-world experiments on a Fetch mobile manipulator across 16 diverse tasks in different environments, which validates the effectiveness of our approach. 

%% file: includes/sections/6-limitations_and_futurework.tex
\section{Limitations and Future Work}\label{sec: limitations}

Our system still has several limitations that can be addressed in future work. First, our execution success relies primarily on object pose estimation using BundleSDF~\cite{Wen2023-bundleSDF}, which is prone to errors under challenging conditions such as large object rotations, suboptimal lighting, and holes in depth images. Currently, we use manual pose verification during task execution to overcome this perception limitation. However, it prevents full autonomy in task execution. In future work, better object pose estimation can significantly benefit our system. For example, one direction is to use strong features in DinoV3~\cite{simeoni2025dinov3} for object pose estimation, or use FoundationStereo~\cite{wen2025foundationstereo} to obtain a better depth input. Another direction is to utilize interactive perception to improve object pose estimation, where the robot can move around and look at the object from different angles until an accurate object pose is obtained.  


Second, our system works in an open-loop manner after the base pose optimization and joint trajectory optimization. While moving to the optimized base location, we limit the velocity of the robot to a very low speed in order to reduce the odometry drift. Higher velocities can amplify odometry errors, leading to inaccurate trajectory for joint trajectory optimization. Future work can consider turning the system into a closed-loop mobile manipulation. For example, our goal is to develop a residual policy with RL that adapts robot actions from trajectory optimization to compensate for errors due to object pose estimation and robot base motion. This policy-driven refinement can add robustness to the framework execution, enabling full autonomy.

Lastly, the trajectory optimization runtime depends on the length of the human demonstration trajectory. If the human demonstration has about 300$\sim$400 frames, the optimization can take 7$\sim$10 seconds. Future work can consider improving the runtime of trajectory optimization through GPU accelerated processing.
